\documentclass[twocolumn,english,compsoc,journal]{IEEEtran}
\usepackage[T1]{fontenc}
\usepackage{babel}
\usepackage{array}
\usepackage{float}
\usepackage{textcomp}
\usepackage{url}
\usepackage{multirow}
\usepackage{amsthm}
\usepackage{amsmath}
\usepackage{graphicx}

\makeatletter

\providecommand{\tabularnewline}{\\}
\floatstyle{ruled}
\newfloat{algorithm}{tbp}{loa}
\providecommand{\algorithmname}{Algorithm}
\floatname{algorithm}{\protect\algorithmname}

 \let\oldforeign@language\foreign@language
 \DeclareRobustCommand{\foreign@language}[1]{%
   \lowercase{\oldforeign@language{#1}}}
\theoremstyle{plain}
\newtheorem{thm}{\protect\theoremname}
\theoremstyle{definition}
\newtheorem{example}[thm]{\protect\examplename}

\usepackage{caption}

\usepackage{xcolor}

\renewcommand\[{\begin{equation}}
\renewcommand\]{\end{equation}}
\renewenvironment{align*}{\align}{\endalign}

\usepackage{bbm}

\makeatother

\usepackage{listings}
\providecommand{\examplename}{Example}
\providecommand{\theoremname}{Theorem}

\begin{document}

\title{High-Speed Tracking with\\
Kernelized Correlation Filters}

\author{João~F.~Henriques, Rui~Caseiro, Pedro~Martins, and~Jorge~Batista\IEEEcompsocitemizethanks{\IEEEcompsocthanksitem The authors are with the Institute of Systems
and Robotics, University of Coimbra.\protect \\
E-mail: \{henriques,ruicaseiro,pedromartins,batista\}@isr.uc.pt

}}

\IEEEaftertitletext{}

\markboth{IEEE Transactions on Pattern Analysis and Machine Intelligence}{Henriques \MakeLowercase{\textit{et al.}}: Kernelized Correlation
Filters}

\IEEEpubid{}

\IEEEtitleabstractindextext{
\begin{abstract}
The core component of most modern trackers is a discriminative classifier,
tasked with distinguishing between the target and the surrounding
environment. To cope with natural image changes, this classifier is
typically trained with translated and scaled sample patches. Such
sets of samples are riddled with redundancies -- any overlapping pixels
are constrained to be the same. Based on this simple observation,
we propose an analytic model for datasets of thousands of translated
patches. By showing that the resulting data matrix is circulant, we
can diagonalize it with the Discrete Fourier Transform, reducing both
storage and computation by several orders of magnitude. Interestingly,
for linear regression our formulation is equivalent to a correlation
filter, used by some of the fastest competitive trackers. For kernel
regression, however, we derive a new Kernelized Correlation Filter
(KCF), that unlike other kernel algorithms has the exact same complexity
as its linear counterpart. Building on it, we also propose a fast
multi-channel extension of linear correlation filters, via a linear
kernel, which we call Dual Correlation Filter (DCF). Both KCF and
DCF outperform top-ranking trackers such as Struck or TLD on a 50
videos benchmark, despite running at hundreds of frames-per-second,
and being implemented in a few lines of code (Algorithm 1). To encourage
further developments, our tracking framework was made open-source.\end{abstract}


}

\maketitle

\IEEEdisplaynontitleabstractindextext{}

\IEEEpeerreviewmaketitle{}

\section{Introduction}

\IEEEPARstart{A}{rguably} one of the biggest breakthroughs in recent
visual tracking research was the widespread adoption of discriminative
learning methods. The task of tracking, a crucial component of many
computer vision systems, can be naturally specified as an online learning
problem \cite{smeulders_visual_2013,yang_recent_2011}. Given an initial
image patch containing the target, the goal is to learn a classifier
to discriminate between its appearance and that of the environment.
This classifier can be evaluated exhaustively at many locations, in
order to detect it in subsequent frames. Of course, each new detection
provides a new image patch that can be used to update the model.

It is tempting to focus on characterizing the object of interest --
the positive samples for the classifier. However, a core tenet of
discriminative methods is to give as much importance, or more, to
the relevant environment -- the negative samples. The most commonly
used negative samples are image patches from different locations and
scales, reflecting the prior knowledge that the classifier will be
evaluated under those conditions.

An extremely challenging factor is the virtually unlimited amount
of negative samples that can be obtained from an image. Due to the
time-sensitive nature of tracking, modern trackers walk a fine line
between incorporating as many samples as possible and keeping computational
demand low. It is common practice to randomly choose only a few samples
each frame \cite{zhang_real-time_2012,kalal_2012,babenko_robust_2011,saffari_-line_2009,hare_struck:_2011}.

Although the reasons for doing so are understandable, we argue that
undersampling negatives is the main factor inhibiting performance
in tracking. In this paper, we develop tools to analytically incorporate
thousands of samples at different relative translations, without iterating
over them explicitly. This is made possible by the discovery that,
in the Fourier domain, some learning algorithms actually become \emph{easier}
as we add \emph{more }samples, if we use a specific model for translations.

These analytical tools, namely circulant matrices, provide a useful
bridge between popular learning algorithms and classical signal processing.
The implication is that we are able to propose a tracker based on
Kernel Ridge Regression \cite{rifkin_regularized_2003} that does
not suffer from the ``curse of kernelization'', which is its larger
asymptotic complexity, and even exhibits lower complexity than unstructured
linear regression. Instead, it can be seen as a kernelized version
of a linear correlation filter, which forms the basis for the fastest
trackers available \cite{bolme_visual_2010,bolme_average_2009}. We
leverage the powerful kernel trick at the same computational complexity
as linear correlation filters. Our framework easily incorporates multiple
feature channels, and by using a linear kernel we show a fast extension
of linear correlation filters to the multi-channel case. 

\begin{figure*}
\begin{centering}
\includegraphics[width=1\textwidth]{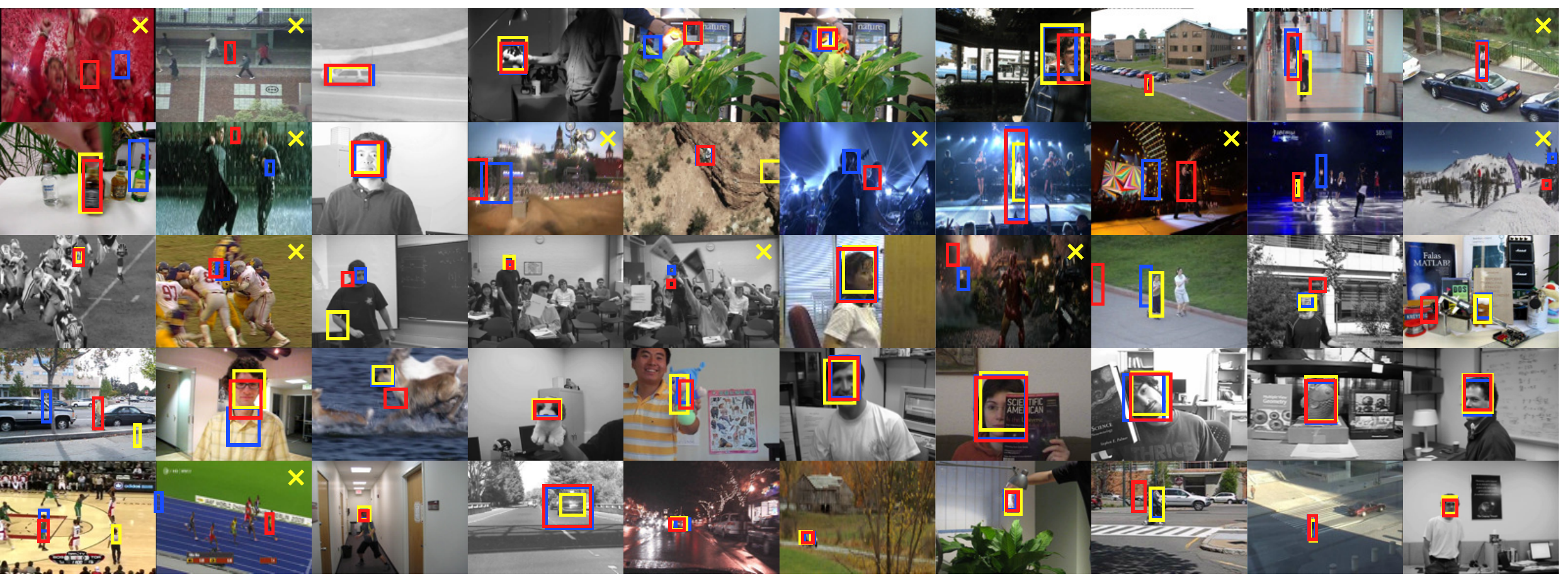}
\par\end{centering}

\begin{raggedleft}
{\footnotesize{}\textcolor{red}{Kernelized Correlation Filter (proposed)} \quad{}
\fcolorbox{yellow}{yellow!50}{TLD} \quad{}
\textcolor{blue}{Struck} \quad{}}
\par\end{raggedleft}{\footnotesize \par}

\protect\caption{Qualitative results for the proposed Kernelized Correlation Filter
(KCF), compared with the top-performing Struck and TLD. Best viewed
on a high-resolution screen. The chosen kernel is Gaussian, on HOG
features. These snapshots were taken at the midpoints of the 50 videos
of a recent benchmark \cite{y_wu_online_2013}. Missing trackers are
denoted by an ``x''. KCF outperforms both Struck and TLD, despite
its minimal implementation and running at 172 FPS (see Algorithm \ref{alg::MATLAB-code-for},
and Table \ref{tab:Summary-of-experimental}).\label{fig:Qualitative-results-for}}
\end{figure*}

\section{Related work}

\subsection{On tracking-by-detection}

A comprehensive review of tracking-by-detection is outside the scope
of this article, but we refer the interested reader to two excellent
and very recent surveys \cite{smeulders_visual_2013,yang_recent_2011}.
The most popular approach is to use a discriminative appearance model
\cite{zhang_real-time_2012,kalal_2012,babenko_robust_2011,saffari_-line_2009}.
It consists of training a classifier online, inspired by statistical
machine learning methods, to predict the presence or absence of the
target in an image patch. This classifier is then tested on many candidate
patches to find the most likely location. Alternatively, the position
can also be predicted directly \cite{hare_struck:_2011}. Regression
with class labels can be seen as classification, so we use the two
terms interchangeably.

We will discuss some relevant trackers before focusing on the literature
that is more directly related to our analytical methods. Canonical
examples of the tracking-by-detection paradigm include those based
on Support Vector Machines (SVM) \cite{avidan_support_2004}, Random
Forest classifiers \cite{saffari_-line_2009}, or boosting variants
\cite{grabner_semi-supervised_2008,babenko_robust_2011}. All the
mentioned algorithms had to be adapted for online learning, in order
to be useful for tracking. Zhang et al. \cite{zhang_real-time_2012}
propose a projection to a fixed random basis, to train a Naive Bayes
classifier, inspired by compressive sensing techniques. Aiming to
predict the target's location directly, instead of its presence in
a given image patch, Hare et al. \cite{hare_struck:_2011} employed
a Structured Output SVM and Gaussian kernels, based on a large number
of image features. Examples of non-discriminative trackers include
the work of Wu et al. \cite{wu_online_2012}, who formulate tracking
as a sequence of image alignment objectives, and of Sevilla-Lara and
Learned-Miller \cite{sevilla_distribution_2012}, who propose a strong
appearance descriptor based on distribution fields. Another discriminative
approach by Kalal et al. \cite{kalal_2012} uses a set of structural
constraints to guide the sampling process of a boosting classifier.
Finally, Bolme et al. \cite{bolme_visual_2010} employ classical signal
processing analysis to derive fast correlation filters. We will discuss
these last two works in more detail shortly.

\subsection{On sample translations and correlation filtering}

Recall that our goal is to learn and detect over translated image
patches efficiently. Unlike our approach, most attempts so far have
focused on trying to weed out irrelevant image patches. On the detection
side, it is possible to use branch-and-bound to find the maximum of
a classifier's response while avoiding unpromising candidate patches
\cite{lampert_beyond_2008}. Unfortunately, in the worst-case the
algorithm may still have to iterate over all patches. A related method
finds the most similar patches of a pair of images efficiently \cite{alexe_exploiting_2011},
but is not directly translated to our setting. Though it does not
preclude an exhaustive search, a notable optimization is to use a
fast but inaccurate classifier to select promising patches, and only
apply the full, slower classifier on those \cite{harzallah_combining_2009,vedaldi_multiple_2009}.

On the training side, Kalal et al. \cite{kalal_2012} propose using
structural constraints to select relevant sample patches from each
new image. This approach is relatively expensive, limiting the features
that can be used, and requires careful tuning of the structural heuristics.
A popular and related method, though it is mainly used in offline
detector learning, is hard-negative mining \cite{felzenszwalb_object_2010}.
It consists of running an initial detector on a pool of images, and
selecting any wrong detections as samples for re-training. Even though
both approaches reduce the number of training samples, a major drawback
is that the candidate patches have to be considered exhaustively,
by running a detector.

The initial motivation for our line of research was the recent success
of correlation filters in tracking \cite{bolme_visual_2010,bolme_average_2009}.
Correlation filters have proved to be competitive with far more complicated
approaches, but using only a fraction of the computational power,
at hundreds of frames-per-second. They take advantage of the fact
that the convolution of two patches (loosely, their dot-product at
different relative translations) is equivalent to an element-wise
product in the Fourier domain. Thus, by formulating their objective
in the Fourier domain, they can specify the desired output of a linear
classifier for several translations, or image shifts, at once.

A Fourier domain approach can be very efficient, and has several decades
of research in signal processing to draw from \cite{gonzalez_digital_2008}.
Unfortunately, it can also be extremely limiting. We would like to
simultaneously leverage more recent advances in computer vision, such
as more powerful features, large-margin classifiers or kernel methods
\cite{DollarPAMI14pyramids,felzenszwalb_object_2010,schoelkopf_learning_2002}.

A few studies go in that direction, and attempt to apply kernel methods
to correlation filters \cite{casasent_analysis_2007,patnaik_fast_2009,jeong2006kernel,xie2005kernel}.
In these works, a distinction must be drawn between two types of objective
functions: those that do not consider the power spectrum or image
translations, such as Synthetic Discriminant Function (SDF) filters
\cite{patnaik_fast_2009,jeong2006kernel}, and those that do, such
as Minimum Average Correlation Energy \cite{mahalanobis1987minimum},
Optimal Trade-Off \cite{xie2005kernel} and Minimum Output Sum of
Squared Error (MOSSE) filters \cite{bolme_visual_2010}. Since the
spatial structure can effectively be ignored, the former are easier
to kernelize, and Kernel SDF filters have been proposed \cite{jeong2006kernel,xie2005kernel,patnaik_fast_2009}.
However, lacking a clearer relationship between translated images,
non-linear kernels and the Fourier domain, applying the kernel trick
to other filters has proven much more difficult \cite{patnaik_fast_2009,casasent_analysis_2007},
with some proposals requiring significantly higher computation times
and imposing strong limits on the number of image shifts that can
be considered \cite{casasent_analysis_2007}.

For us, this hinted that a deeper connection between translated image
patches and training algorithms was needed, in order to overcome the
limitations of direct Fourier domain formulations.

\subsection{Subsequent work}

Since the initial version of this work \cite{henriques2012circulant},
an interesting time-domain variant of the proposed cyclic shift model
has been used very successfully for video event retrieval \cite{revaud_event_2013}.
Generalizations of linear correlation filters to multiple channels
have also been proposed \cite{henriques2013beyond,galoogahi_multi-channel_2013,boddeti_correlation_2013},
some of which building on our initial work. This allows them to leverage
more modern features (e.g. Histogram of Oriented Gradients -- HOG).
A generalization to other linear algorithms, such as Support Vector
Regression, was also proposed \cite{henriques2013beyond}. We must
point out that all of these works target off-line training, and thus
rely on slower solvers \cite{henriques2013beyond,galoogahi_multi-channel_2013,boddeti_correlation_2013}.
In contrast, we focus on fast element-wise operations, which are more
suitable for real-time tracking, even with the kernel trick.

\section{Contributions}

A preliminary version of this work was presented earlier \cite{henriques2012circulant}.
It demonstrated, for the first time, the connection between Ridge
Regression with cyclically shifted samples and classical correlation
filters. This enabled fast learning with $\mathcal{O}\left(n\log n\right)$
Fast Fourier Transforms instead of expensive matrix algebra. The first
Kernelized Correlation Filter was also proposed, though limited to
a single channel. Additionally, it proposed closed-form solutions
to compute kernels at all cyclic shifts. These carried the same $\mathcal{O}\left(n\log n\right)$
computational cost, and were derived for radial basis and dot-product
kernels.

The present work adds to the initial version in significant ways.
All the original results were re-derived using a much simpler diagonalization
technique (Sections \ref{sec:Building-blocks}-\ref{sec:Fast-kernel-correlation}).
We extend the original work to deal with multiple channels, which
allows the use of state-of-the-art features that give an important
boost to performance (Section \ref{sec:Multiple-channels}). Considerable
new analysis and intuitive explanations are added to the initial results.
We also extend the original experiments from 12 to 50 videos, and
add a new variant of the Kernelized Correlation Filter (KCF) tracker
based on Histogram of Oriented Gradients (HOG) features instead of
raw pixels. Via a linear kernel, we additionally propose a linear
multi-channel filter with very low computational complexity, that
almost matches the performance of non-linear kernels. We name it Dual
Correlation Filter (DCF), and show how it is related to a set of recent,
more expensive multi-channel filters \cite{henriques2013beyond}.
Experimentally, we demonstrate that the KCF already performs better
than a linear filter, without any feature extraction. With HOG features,
both the linear DCF and non-linear KCF outperform by a large margin
top-ranking trackers, such as Struck \cite{hare_struck:_2011} or
Track-Learn-Detect (TLD) \cite{kalal_2012}, while comfortably running
at hundreds of frames-per-second.

\section{Building blocks\label{sec:Building-blocks}}

In this section, we propose an analytical model for image patches
extracted at different translations, and work out the impact on a
linear regression algorithm. We will show a natural underlying connection
to classical correlation filters. The tools we develop will allow
us to study more complicated algorithms in Sections \ref{sec:Non-linear-regression}-\ref{sec:Multiple-channels}.

\subsection{Linear regression}

We will focus on Ridge Regression, since it admits a simple closed-form
solution, and can achieve performance that is close to more sophisticated
methods, such as Support Vector Machines \cite{rifkin_regularized_2003}.
The goal of training is to find a function $f(\mathbf{z})=\mathbf{w}^{T}\mathbf{z}$
that minimizes the squared error over samples $\mathbf{x}_{i}$ and
their regression targets $y_{i}$,

\[
\underset{\mathbf{w}}{\min}\,\sum_{i}\left(f(\mathbf{x}_{i})-y_{i}\right)^{2}+\lambda\left\Vert \mathbf{w}\right\Vert ^{2}.
\]

The $\lambda$ is a regularization parameter that controls overfitting,
as in the SVM. As mentioned earlier, the minimizer has a closed-form,
which is given by \cite{rifkin_regularized_2003}

\begin{equation}
\mathbf{w}=\left(X^{T}X+\lambda I\right)^{-1}X^{T}\mathbf{y}.\label{eq:real_regression}
\end{equation}
where the data matrix $X$ has one sample per row $\mathbf{x}_{i}$,
and each element of $\mathbf{y}$ is a regression target $y_{i}$.
$I$ is an identity matrix.

Starting in Section \ref{sub:Putting-it-all}, we will have to work
in the Fourier domain, where quantities are usually complex-valued.
They are not harder to deal with, as long as we use the complex version
of Eq. \ref{eq:real_regression} instead,

\begin{equation}
\mathbf{w}=\left(X^{H}X+\lambda I\right)^{-1}X^{H}\mathbf{y},\label{eq:linear_ridge_regression}
\end{equation}
where $X^{H}$ is the Hermitian transpose, i.e., $X^{H}=\left(X^{*}\right)^{T}$,
and $X^{*}$ is the complex-conjugate of $X$. For real numbers, Eq.
\ref{eq:linear_ridge_regression} reduces to Eq. \ref{eq:real_regression}.

In general, a large system of linear equations must be solved to compute
the solution, which can become prohibitive in a real-time setting.
Over the next paragraphs we will see a special case of $\mathbf{x}_{i}$
that bypasses this limitation.

\pagebreak{}

\begin{figure}
{\setlength{\tabcolsep}{0em}

\begin{centering}
\begin{tabular}{>{\centering}p{0.2\columnwidth}>{\centering}p{0.2\columnwidth}>{\centering}p{0.2\columnwidth}>{\centering}p{0.2\columnwidth}>{\centering}p{0.2\columnwidth}}
\includegraphics[width=0.19\columnwidth]{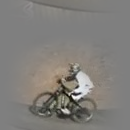} & \includegraphics[width=0.19\columnwidth]{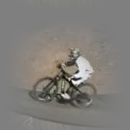} & \includegraphics[width=0.19\columnwidth]{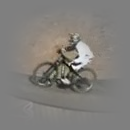} & \includegraphics[width=0.19\columnwidth]{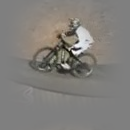} & \includegraphics[width=0.19\columnwidth]{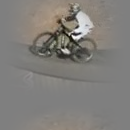}\tabularnewline
{\small{}+30} & {\small{}+15} & {\small{}Base sample} & {\small{}-15} & {\small{}-30}\tabularnewline
\end{tabular}
\par\end{centering}

}

\protect\caption{Examples of vertical cyclic shifts of a base sample. Our Fourier domain
formulation allows us to train a tracker with \emph{all} possible
cyclic shifts of a base sample, both vertical and horizontal, without
iterating them explicitly. Artifacts from the wrapped-around edges
can be seen (top of the left-most image), but are mitigated by the
cosine window and padding.\label{fig:cyclic_shifts}}
\end{figure}

\subsection{Cyclic shifts}

For notational simplicity, we will focus on single-channel, one-dimensional
signals. These results generalize to multi-channel, two-dimensional
images in a straightforward way (Section \ref{sec:Multiple-channels}).

Consider an $n\times1$ vector representing a patch with the object
of interest, denoted $\mathbf{x}$. We will refer to it as the \emph{base
sample}. Our goal is to train a classifier with both the base sample
(a positive example) and several virtual samples obtained by translating
it (which serve as negative examples). We can model one-dimensional
translations of this vector by a \emph{cyclic shift operator}, which
is the permutation matrix

\[
P=\left[\begin{array}{ccccc}
0 & 0 & 0 & \cdots & 1\\
1 & 0 & 0 & \cdots & 0\\
0 & 1 & 0 & \cdots & 0\\
\vdots & \vdots & \ddots & \ddots & \vdots\\
0 & 0 & \cdots & 1 & 0
\end{array}\right].
\]

\begin{figure}
\begin{centering}
\includegraphics[width=0.95\columnwidth]{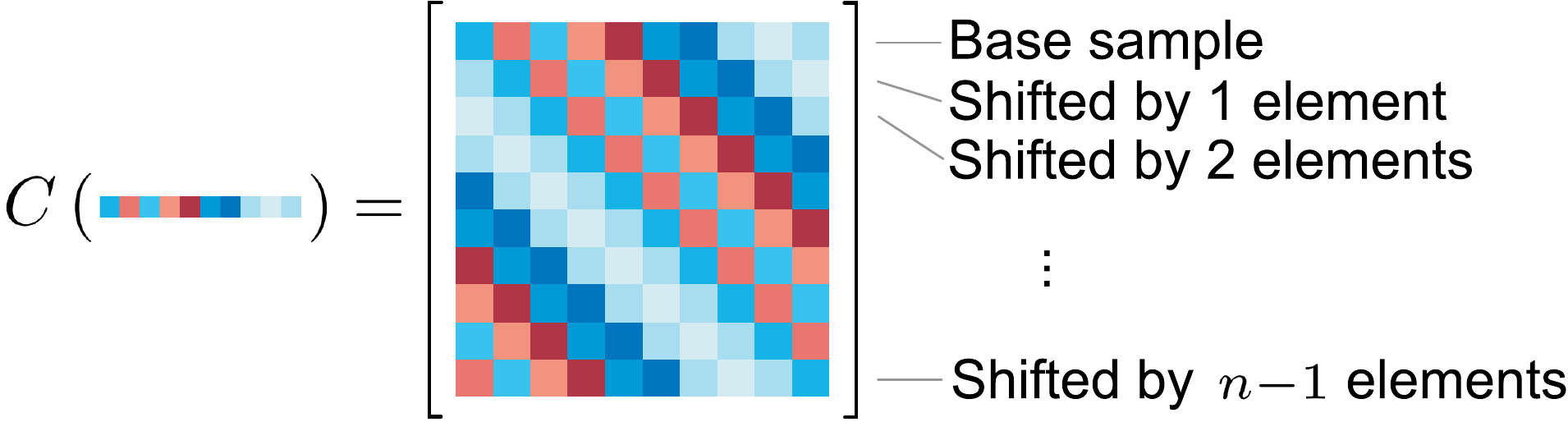}
\par\end{centering}

\protect\caption{Illustration of a circulant matrix. The rows are cyclic shifts of
a vector image, or its translations in 1D. The same properties carry
over to circulant matrices containing 2D images.\label{fig:circ-illustration}}
\end{figure}

The product $P\mathbf{x}=\left[x_{n},x_{1},x_{2},\ldots,x_{n-1}\right]^{T}$
shifts $\mathbf{x}$ by one element, modeling a small translation.
We can chain $u$ shifts to achieve a larger translation by using
the matrix power $P^{u}\mathbf{x}$. A negative $u$ will shift in
the reverse direction. A 1D signal translated horizontally with this
model is illustrated in Fig. \ref{fig:circ-illustration}, and an
example for a 2D image is shown in Fig. \ref{fig:cyclic_shifts}.

The attentive reader will notice that the last element wraps around,
inducing some distortion relative to a true translation. However,
this undesirable property can be mitigated by appropriate padding
and windowing (Section \ref{sub:Implementation-details}). The fact
that a large percentage of the elements of a signal are still modeled
correctly, even for relatively large translations (see Fig. \ref{fig:cyclic_shifts}),
explains the observation that cyclic shifts work well in practice.

Due to the cyclic property, we get the same signal $\mathbf{x}$ periodically
every $n$ shifts. This means that the full set of shifted signals
is obtained with

\begin{equation}
\left\{ P^{u}\mathbf{x}\,|\, u=0,\ldots n-1\right\} .\label{eq:shifted_samples}
\end{equation}

Again due to the cyclic property, we can equivalently view the first
half of this set as shifts in the positive direction, and the second
half as shifts in the negative direction.

\subsection{Circulant matrices}

To compute a regression with shifted samples, we can use the set of
Eq. \ref{eq:shifted_samples} as the rows of a data matrix $X$:

\begin{equation}
X=C(\mathbf{x})=\left[\begin{array}{ccccc}
x_{1} & x_{2} & x_{3} & \cdots & x_{n}\\
x_{n} & x_{1} & x_{2} & \cdots & x_{n-1}\\
x_{n-1} & x_{n} & x_{1} & \cdots & x_{n-2}\\
\vdots & \vdots & \vdots & \ddots & \vdots\\
x_{2} & x_{3} & x_{4} & \cdots & x_{1}
\end{array}\right].\label{eq:X}
\end{equation}

An illustration of the resulting pattern is given in Fig. \ref{fig:circ-illustration}.
What we have just arrived at is a \emph{circulant} matrix, which has
several intriguing properties \cite{gray_toeplitz_2006,davis1994circulant}.
Notice that the pattern is deterministic, and fully specified by the
generating vector $\mathbf{x}$, which is the first row.

What is perhaps most amazing and useful is the fact that \emph{all}
circulant matrices are made diagonal by the Discrete Fourier Transform
(DFT), regardless of the generating vector $\mathbf{x}$ \cite{gray_toeplitz_2006}.
This can be expressed as

\begin{equation}
X=F\,\textrm{diag}\left(\hat{\mathbf{x}}\right)\, F^{H},\label{eq:eigen}
\end{equation}
where $F$ is a constant matrix that does not depend on $\mathbf{x}$,
and $\hat{\mathbf{x}}$ denotes the DFT of the generating vector,
$\hat{\mathbf{x}}=\mathcal{F}\left(\mathbf{x}\right)$. From now on,
we will always use a hat $\hat{\,}$ as shorthand for the DFT of a
vector.

The constant matrix $F$ is known as the \emph{DFT matrix}, and is
the unique matrix that computes the DFT of any input vector, as $\mathcal{F}\left(\mathbf{z}\right)=\sqrt{n}F\mathbf{z}$.
This is possible because the DFT is a linear operation.

Eq. \ref{eq:eigen} expresses the eigendecomposition of a general
circulant matrix. The shared, deterministic eigenvectors $F$ lie
at the root of many uncommon features, such as commutativity or closed-form
inversion.

\subsection{Putting it all together\label{sub:Putting-it-all}}

We can now apply this new knowledge to simplify the linear regression
in Eq. \ref{eq:linear_ridge_regression}, when the training data is
composed of cyclic shifts. Being able to work solely with diagonal
matrices is very appealing, because all operations can be done element-wise
on their diagonal elements.

Take the term $X^{H}X$, which can be seen as a non-centered covariance
matrix. Replacing Eq. \ref{eq:eigen} in it,

\[
X^{H}X=F\,\textrm{diag}\left(\hat{\mathbf{x}}^{*}\right)\, F^{H}F\,\textrm{diag}\left(\hat{\mathbf{x}}\right)\, F^{H}.
\]

Since diagonal matrices are symmetric, taking the Hermitian transpose
only left behind a complex-conjugate, $\hat{\mathbf{x}}^{*}$. Additionally,
we can eliminate the factor $F^{H}F=I$. This property is the unitarity
of $F$ and can be canceled out in many expressions. We are left with

\[
X^{H}X=F\,\textrm{diag}\left(\hat{\mathbf{x}}^{*}\right)\,\textrm{diag}\left(\hat{\mathbf{x}}\right)\, F^{H}.
\]

Because operations on diagonal matrices are element-wise, we can define
the element-wise product as $\odot$ and obtain

\begin{equation}
X^{H}X=F\,\textrm{diag}\left(\hat{\mathbf{x}}^{*}\odot\hat{\mathbf{x}}\right)\, F^{H}.\label{eq:non-centered-cov}
\end{equation}

An interesting aspect is that the vector in brackets is known as the
\emph{auto-correlation} of the signal $\mathbf{x}$ (in the Fourier
domain, also known as the power spectrum \cite{gonzalez_digital_2008}).
In classical signal processing, it contains the variance of a time-varying
process for different time lags, or in our case, space.

The above steps summarize the general approach taken in diagonalizing
expressions with circulant matrices. Applying them recursively to
the full expression for linear regression (Eq. \ref{eq:linear_ridge_regression}),
we can put most quantities inside the diagonal,

\[
\hat{\mathbf{w}}=\textrm{diag}\left(\frac{\hat{\mathbf{x}}^{*}}{\hat{\mathbf{x}}^{*}\odot\hat{\mathbf{x}}+\lambda}\right)\hat{\mathbf{y}},
\]
or better yet,

\begin{equation}
\hat{\mathbf{w}}=\frac{\hat{\mathbf{x}}^{*}\odot\hat{\mathbf{y}}}{\hat{\mathbf{x}}^{*}\odot\hat{\mathbf{x}}+\lambda}.\label{eq:mosse}
\end{equation}

The fraction denotes element-wise division. We can easily recover
$\mathbf{w}$ in the spatial domain with the Inverse DFT, which has
the same cost as a forward DFT. The detailed steps of the recursive
diagonalization that yields Eq. \ref{eq:mosse} are given in Appendix
\ref{sub:appendix-linear}.

At this point we just found an unexpected formula from classical signal
processing -- the solution is a regularized correlation filter \cite{bolme_visual_2010,gonzalez_digital_2008}.

Before exploring this relation further, we must highlight the computational
efficiency of Eq. \ref{eq:mosse}, compared to the prevalent method
of extracting patches explicitly and solving a general regression
problem. For example, Ridge Regression has a cost of $\mathcal{O}\left(n^{3}\right)$,
bound by the matrix inversion and products%
\footnote{We remark that the complexity of training algorithms is usually reported
in terms of the number of samples $n$, disregarding the number of
features $m$. Since in our case $m=n$ ($X$ is square), we conflate
the two quantities. For comparison, the fastest SVM solvers have ``linear''
complexity in the samples $\mathcal{O}\left(mn\right)$, but under
the same conditions $m=n$ would actually exhibit quadratic complexity,
$\mathcal{O}\left(n^{2}\right)$.%
}. On the other hand, all operations in Eq. \ref{eq:mosse} are element-wise
($\mathcal{O}\left(n\right)$), except for the DFT, which bounds the
cost at a nearly-linear $\mathcal{O}\left(n\log n\right)$. For typical
data sizes, this reduces storage and computation by several orders
of magnitude.

\subsection{Relationship to correlation filters}

Correlation filters have been a part of signal processing since the
80's, with solutions to a myriad of objective functions in the Fourier
domain \cite{gonzalez_digital_2008,mahalanobis1987minimum}. Recently,
they made a reappearance as MOSSE filters \cite{bolme_visual_2010},
which have shown remarkable performance in tracking, despite their
simplicity and high FPS rate.

The solution to these filters looks like Eq. \ref{eq:mosse} (see
Appendix A.2), but with two crucial differences. First, MOSSE filters
are derived from an objective function specifically formulated in
the Fourier domain. Second, the $\lambda$ regularizer is added in
an ad-hoc way, to avoid division-by-zero. The derivation we showed
above adds considerable insight, by specifying the starting point
as Ridge Regression with cyclic shifts, and arriving at the same solution.

Circulant matrices allow us to enrich the toolset put forward by classical
signal processing and modern correlation filters, and apply the Fourier
trick to new algorithms. Over the next section we will see one such
instance, in training non-linear filters.

\section{Non-linear regression\label{sec:Non-linear-regression}}

One way to allow more powerful, non-linear regression functions $f(\mathbf{z})$
is with the \emph{``}kernel trick\emph{''} \cite{schoelkopf_learning_2002}.
The most attractive quality is that the optimization problem is still
linear, albeit in a different set of variables (the \emph{dual }space).
On the downside, evaluating $f(\mathbf{z})$ typically grows in complexity
with the number of samples.

Using our new analysis tools, however, we will show that it is possible
to overcome this limitation, and obtain non-linear filters that are
\emph{as fast as linear correlation filters}, both to train and evaluate.

\subsection{Kernel trick -- brief overview}

This section will briefly review the kernel trick, and define the
relevant notation.

Mapping the inputs of a linear problem to a non-linear feature-space
$\varphi(\mathbf{x})$ with the kernel trick consists of:

\smallskip{}

\begin{enumerate}
\item Expressing the solution $\mathbf{w}$ as a linear combination of the
samples: 
\[
\mathbf{w}=\sum_{i}\alpha_{i}\varphi(\mathbf{x}_{i})
\]
The variables under optimization are thus $\boldsymbol{\alpha}$,
instead of $\mathbf{w}$. This alternative representation $\boldsymbol{\alpha}$
is said to be in the \emph{dual space}, as opposed to the \emph{primal
space} $\mathbf{w}$ (Representer Theorem \cite[p. 89]{schoelkopf_learning_2002}).
\end{enumerate}
\smallskip{}

\begin{enumerate}
\item Writing the algorithm in terms of dot-products $\varphi^{T}(\mathbf{x})\varphi(\mathbf{x}')=\kappa(\mathbf{x},\mathbf{x}')$,
which are computed using the kernel function $\kappa$ (e.g., Gaussian
or Polynomial).
\end{enumerate}
\smallskip{}

The dot-products between all pairs of samples are usually stored in
a $n\times n$ kernel matrix $K$, with elements

\begin{equation}
K_{ij}=\kappa(\mathbf{x}_{i},\mathbf{x}_{j}).\label{eq:k_matrix}
\end{equation}

The power of the kernel trick comes from the implicit use of a high-dimensional
feature space $\varphi(\mathbf{x})$, without ever instantiating a
vector in that space. Unfortunately, this is also its greatest weakness,
since the regression function's complexity grows with the number of
samples,

\begin{equation}
f(\mathbf{z})=\mathbf{w}^{T}\mathbf{z}=\sum_{i=1}^{n}\alpha_{i}\kappa(\mathbf{z},\mathbf{x}_{i}).\label{eq:k_classifier}
\end{equation}

In the coming sections we will show how most drawbacks of the kernel
trick can be avoided, assuming circulant data.

\subsection{Fast kernel regression\label{sub:Fast-kernel-regression}}

The solution to the kernelized version of Ridge Regression is given
by \cite{rifkin_regularized_2003}

\begin{equation}
\boldsymbol{\alpha}=\left(K+\lambda I\right)^{-1}\mathbf{y},\label{eq:krr}
\end{equation}
where $K$ is the kernel matrix and $\boldsymbol{\alpha}$ is the
vector of coefficients $\alpha_{i}$, that represent the solution
in the dual space.

Now, if we can prove that $K$ is circulant for datasets of cyclic
shifts, we can diagonalize Eq. \ref{eq:krr} and obtain a fast solution
as for the linear case. This would seem to be intuitively true, but
does not hold in general. The arbitrary non-linear mapping $\varphi(\mathbf{x})$
gives us no guarantee of preserving any sort of structure. However,
we can impose one condition that will allow $K$ to be circulant.
It turns out to be fairly broad, and apply to most useful kernels.\medskip{}

\begin{thm}
Given circulant data $C\left(\mathbf{x}\right)$, the corresponding
kernel matrix $K$ is circulant if the kernel function satisfies $\kappa(\mathbf{x},\mathbf{x}')=\kappa(M\mathbf{x},M\mathbf{x}')$,
for any permutation matrix $M$.\label{thm:circulant-K}
\end{thm}
\medskip{}

For a proof, see Appendix \ref{sub:Proof-of-Theorem}. What this means
is that, for a kernel to preserve the circulant structure, it must
treat all dimensions of the data equally. Fortunately, this includes
most useful kernels.

\medskip{}

\begin{example}
\emph{The following kernels satisfy Theorem \ref{thm:circulant-K}:}
\begin{itemize}
\item \emph{Radial Basis Function kernels -- e.g., Gaussian.}
\item \emph{Dot-product kernels -- e.g., linear, polynomial.}
\item \emph{Additive kernels -- e.g., intersection, $\chi^{2}$ and Hellinger
kernels \cite{vedaldi_efficient_2011}.}
\item \emph{Exponentiated additive kernels.}
\end{itemize}
\end{example}
\medskip{}
Checking this fact is easy, since reordering the dimensions of $\mathbf{x}$
and $\mathbf{x}'$ simultaneously does not change $\kappa(\mathbf{x},\mathbf{x}')$
for these kernels. This applies to any kernel that combines dimensions
through a commutative operation, such as sum, product, min and max.

\medskip{}

Knowing which kernels we can use to make $K$ circulant, it is possible
to diagonalize Eq. \ref{eq:krr} as in the linear case, obtaining

\begin{equation}
\hat{\boldsymbol{\alpha}}=\frac{\hat{\mathbf{y}}}{\hat{\mathbf{k}}{}^{\mathbf{xx}}+\lambda},\label{eq:fast_krr}
\end{equation}
where $\mathbf{k}^{\mathbf{xx}}$ is the first row of the kernel matrix
$K=C(\mathbf{k}^{\mathbf{xx}})$, and again a hat $\hat{\,}$ denotes
the DFT of a vector. A detailed derivation is in Appendix \ref{sub:appendix-Kernel-Ridge-Regression}.

To better understand the role of $\mathbf{k}^{\mathbf{xx}}$, we found
it useful to define a more general \emph{kernel correlation}. The
kernel correlation of two arbitrary vectors, $\mathbf{x}$ and $\mathbf{x}'$,
is the vector\emph{ }$\mathbf{k}^{\mathbf{xx}'}$ with elements

\begin{equation}
k_{i}^{\mathbf{xx}'}=\kappa(\mathbf{x}',\, P^{i-1}\mathbf{x}).\label{eq:k_corr}
\end{equation}
In words, it contains the kernel evaluated for different relative
shifts of the two arguments. Then $\hat{\mathbf{k}}^{\mathbf{xx}}$
is the kernel correlation of $\mathbf{x}$ with itself, in the Fourier
domain. We can refer to it as the \emph{kernel auto-correlation},
in analogy with the linear case.

This analogy can be taken further. Since a kernel is equivalent to
a dot-product in a high-dimensional space $\varphi(\cdot)$, another
way to view Eq. \ref{eq:k_corr} is

\[
k_{i}^{\mathbf{xx}'}=\varphi^{T}(\mathbf{x}')\varphi(P^{i-1}\mathbf{x}),
\]
which is the cross-correlation of $\mathbf{x}$ and $\mathbf{x}'$
in the high-dimensional space $\varphi(\cdot)$.

Notice how we only need to compute and operate on the kernel auto-correlation,
an $n\times1$ vector, which grows linearly with the number of samples.
This is contrary to the conventional wisdom on kernel methods, which
requires computing an $n\times n$ kernel matrix, scaling quadratically
with the samples. Our knowledge of the exact structure of $K$ allowed
us to do better than a generic algorithm.

Finding the optimal $\boldsymbol{\alpha}$ is not the only problem
that can be accelerated, due to the ubiquity of translated patches
in a tracking-by-detection setting. Over the next paragraphs we will
investigate the effect of the cyclic shift model on the detection
phase, and even in computing kernel correlations.

\subsection{Fast detection\label{sub:Fast-detection}}

It is rarely the case that we want to evaluate the regression function
$f(\mathbf{z})$ for one image patch in isolation. To detect the object
of interest, we typically wish to evaluate $f(\mathbf{z})$ on several
image locations, i.e., for several candidate patches. These patches
can be modeled by cyclic shifts.

Denote by $K^{\mathbf{z}}$ the (asymmetric) kernel matrix between
all training samples and all candidate patches. Since the samples
and patches are cyclic shifts of base sample $\mathbf{x}$ and base
patch $\mathbf{z}$, respectively, each element of $K^{\mathbf{z}}$
is given by $\kappa(P^{i-1}\mathbf{z},\, P^{j-1}\mathbf{x})$. It
is easy to verify that this kernel matrix satisfies Theorem \ref{thm:circulant-K},
and is circulant for appropriate kernels.

Similarly to Section \ref{sub:Fast-kernel-regression}, we only need
the first row to define the kernel matrix:

\[
K^{\mathbf{z}}=C(\mathbf{k}^{\mathbf{xz}}),
\]
where $\mathbf{k}^{\mathbf{xz}}$ is the \emph{kernel correlation}
of $\mathbf{x}$ and $\mathbf{z}$, as defined before.

From Eq. \ref{eq:k_classifier}, we can compute the regression function
for all candidate patches with

\begin{equation}
\mathbf{f}(\mathbf{z})=\left(K^{\mathbf{z}}\right)^{T}\boldsymbol{\alpha}.\label{eq:detection}
\end{equation}

Notice that $\mathbf{f}(\mathbf{z})$ is a vector, containing the
output for \emph{all} cyclic shifts of $\mathbf{z}$, i.e., the full
detection response. To compute Eq. \ref{eq:detection} efficiently,
we diagonalize it to obtain

\begin{equation}
\hat{\mathbf{f}}(\mathbf{z})=\hat{\mathbf{k}}^{\mathbf{xz}}\odot\hat{\boldsymbol{\alpha}}.\label{eq:fast-detection}
\end{equation}

Intuitively, evaluating $f(\mathbf{z})$ at all locations can be seen
as a spatial filtering operation over the kernel values $\mathbf{k}^{\mathbf{xz}}$.
Each $f(\mathbf{z})$ is a linear combination of the neighboring kernel
values from $\mathbf{k}^{\mathbf{xz}}$, weighted by the learned coefficients
$\boldsymbol{\alpha}$. Since this is a filtering operation, it can
be formulated more efficiently in the Fourier domain.

\section{Fast kernel correlation\label{sec:Fast-kernel-correlation}}

Even though we have found faster algorithms for training and detection,
they still rely on computing one kernel correlation each ($\mathbf{k}^{\mathbf{xx}}$
and $\mathbf{k}^{\mathbf{xz}}$, respectively). Recall that kernel
correlation consists of computing the kernel for all relative shifts
of two input vectors. This represents the last standing computational
bottleneck, as a naive evaluation of $n$ kernels for signals of size
$n$ will have quadratic complexity. However, using the cyclic shift
model will allow us to efficiently exploit the redundancies in this
expensive computation.

\subsection{Dot-product and polynomial kernels\label{sub:Dot-product-kernels}}

Dot-product kernels have the form $\kappa(\mathbf{x},\mathbf{x}')=g(\mathbf{x}^{T}\mathbf{x}')$,
for some function $g$. Then, $\mathbf{k}^{\mathbf{xx}'}$ has elements

\begin{equation}
k_{i}^{\mathbf{xx}'}=\kappa(\mathbf{x}',\, P^{i-1}\mathbf{x})=g\negthinspace\left(\mathbf{x}'^{T}P^{i-1}\mathbf{x}\right).\label{eq:dot-prod-kernel-elementwise}
\end{equation}

Let $g$ also work element-wise on any input vector. This way we can
write Eq. \ref{eq:dot-prod-kernel-elementwise} in vector form

\[
\mathbf{k}^{\mathbf{xx}'}=g\negthinspace\left(C\negthinspace(\mathbf{x})\,\mathbf{x}'\right).
\]

This makes it an easy target for diagonalization, yielding

\begin{equation}
\mathbf{k}^{\mathbf{xx}'}=g\negthinspace\left(\mathcal{F}^{-1}\left(\hat{\mathbf{x}}^{*}\odot\hat{\mathbf{x}}'\right)\right),\label{eq:dot-product-kernel}
\end{equation}
where $\mathcal{F}^{-1}$ denotes the Inverse DFT.

In particular, for a polynomial kernel $\kappa(\mathbf{x},\mathbf{x}')=\left(\mathbf{x}^{T}\mathbf{x}'+a\right)^{b}$,

\[
\mathbf{k}^{\mathbf{xx}'}=\left(\mathcal{F}^{-1}\left(\hat{\mathbf{x}}^{*}\odot\hat{\mathbf{x}}'\right)+a\right)^{b}.
\]

Then, computing the kernel correlation for these particular kernels
can be done using only a few DFT/IDFT and element-wise operations,
in $\mathcal{O}\left(n\log n\right)$ time.

\subsection{Radial Basis Function and Gaussian kernels\label{sub:Radial-Basis-Function}}

RBF kernels have the form $\kappa(\mathbf{x},\mathbf{x}')=h(\left\Vert \mathbf{x}-\mathbf{x}'\right\Vert ^{2})$,
for some function $h$. The elements of $\mathbf{k}^{\mathbf{xx}'}$
are

\[
k_{i}^{\mathbf{xx}'}=\kappa(\mathbf{x}',\, P^{i-1}\mathbf{x})=h\negthinspace\left(\left\Vert \mathbf{x}'-P^{i-1}\mathbf{x}\right\Vert ^{2}\right)
\]

We will show (Eq. \ref{eq:rbf}) that this is actually a special case
of a dot-product kernel. We only have to expand the norm,

\begin{equation}
k_{i}^{\mathbf{xx}'}=h\negthinspace\left(\left\Vert \mathbf{x}\right\Vert ^{2}+\left\Vert \mathbf{x}'\right\Vert ^{2}-2\mathbf{x}'^{T}P^{i-1}\mathbf{x}\right).\label{eq:rbf_element}
\end{equation}

The permutation $P^{i-1}$ does not affect the norm of $\mathbf{x}$
due to Parseval's Theorem \cite{gonzalez_digital_2008}. Since $\left\Vert \mathbf{x}\right\Vert ^{2}$
and $\left\Vert \mathbf{x}'\right\Vert ^{2}$ are constant w.r.t.
$i$, Eq. \ref{eq:rbf_element} has the same form as a dot-product
kernel (Eq. \ref{eq:dot-prod-kernel-elementwise}). Leveraging the
result from the previous section,

\begin{equation}
\mathbf{k}^{\mathbf{xx}'}=h\negthinspace\left(\left\Vert \mathbf{x}\right\Vert ^{2}+\left\Vert \mathbf{x}'\right\Vert ^{2}-2\mathcal{F}^{-1}\left(\hat{\mathbf{x}}^{*}\odot\hat{\mathbf{x}}'\right)\right).\label{eq:rbf}
\end{equation}

As a particularly useful special case, for a Gaussian kernel $\kappa(\mathbf{x},\mathbf{x}')=\exp\left(-\frac{1}{\sigma^{2}}\left\Vert \mathbf{x}-\mathbf{x}'\right\Vert ^{2}\right)$
we get

\begin{equation}
\mathbf{k}^{\mathbf{xx}'}=\exp\left(-\frac{1}{\sigma^{2}}\left(\left\Vert \mathbf{x}\right\Vert ^{2}+\left\Vert \mathbf{x}'\right\Vert ^{2}-2\mathcal{F}^{-1}\left(\hat{\mathbf{x}}^{*}\odot\hat{\mathbf{x}}'\right)\right)\right).\label{eq:gauss}
\end{equation}

As before, we can compute the full kernel correlation in only $\mathcal{O}\left(n\log n\right)$
time.

\subsection{Other kernels}

The approach from the preceding two sections depends on the kernel
value being unchanged by unitary transformations, such as the DFT.
This does not hold in general for other kernels, e.g. intersection
kernel. We can still use the fast training and detection results (Sections
\ref{sub:Fast-kernel-regression} and \ref{sub:Fast-detection}),
but kernel correlation must be evaluated by a more expensive sliding
window method.

\section{Multiple channels\label{sec:Multiple-channels}}

In this section, we will see that working in the dual has the advantage
of allowing multiple channels (such as the orientation bins of a HOG
descriptor \cite{felzenszwalb_object_2010}) by simply summing over
them in the Fourier domain. This characteristic extends to the linear
case, simplifying the recently-proposed multi-channel correlation
filters \cite{henriques2013beyond,galoogahi_multi-channel_2013,boddeti_correlation_2013}
considerably, under specific conditions.

\subsection{General case\label{sub:multi-channel-Non-linear-kernels}}

To deal with multiple channels, in this section we will assume that
a vector $\mathbf{x}$ concatenates the individual vectors for $C$
channels (e.g. 31 gradient orientation bins for a HOG variant \cite{felzenszwalb_object_2010}),
as $\mathbf{x}=\left[\mathbf{x}_{1},\ldots,\mathbf{x}_{C}\right]$.

Notice that all kernels studied in Section \ref{sec:Fast-kernel-correlation}
are based on either dot-products or norms of the arguments. A dot-product
can be computed by simply summing the individual dot-products for
each channel. By linearity of the DFT, this allows us to sum the result
for each channel in the Fourier domain. As a concrete example, we
can apply this reasoning to the Gaussian kernel, obtaining the multi-channel
analogue of Eq. \ref{eq:gauss},

\begin{equation}
\mathbf{k}^{\mathbf{xx}'}=\exp\left(-\frac{1}{\sigma^{2}}\left(\left\Vert \mathbf{x}\right\Vert ^{2}+\left\Vert \mathbf{x}'\right\Vert ^{2}-2\mathcal{F}^{-1}\left({\textstyle \sum_{c}}\,\hat{\mathbf{x}}_{c}^{*}\odot\hat{\mathbf{x}}'{}_{c}\right)\right)\right).\label{eq:gauss-multi}
\end{equation}

It is worth emphasizing that the integration of multiple channels
does not result in a more difficult inference problem -- we merely
have to sum over the channels when computing kernel correlation.

\subsection{Linear kernel}

For a linear kernel $\kappa(\mathbf{x},\mathbf{x}')=\mathbf{x}^{T}\mathbf{x}'$,
the multi-channel extension from the previous section simply yields

\begin{equation}
\mathbf{k}^{\mathbf{xx}'}=\mathcal{F}^{-1}\left({\textstyle \sum_{c}}\,\hat{\mathbf{x}}_{c}^{*}\odot\hat{\mathbf{x}}'{}_{c}\right).\label{eq:multi-channel-linear-kernel}
\end{equation}

We named it the Dual Correlation Filter (DCF). This filter is linear,
but trained in the dual space $\boldsymbol{\alpha}$. We will discuss
the advantages over other multi-channel filters shortly.

A recent extension of linear correlation filters to multiple channels
was discovered independently by three groups \cite{henriques2013beyond,galoogahi_multi-channel_2013,boddeti_correlation_2013}.
They allow much faster training times than unstructured algorithms,
by decomposing the problem into one linear system for each DFT frequency,
in the case of Ridge Regression. Henriques et al. \cite{henriques2013beyond}
additionally generalize the decomposition to other training algorithms.

However, Eq. \ref{eq:multi-channel-linear-kernel} suggests that,
by working in the dual with a linear kernel, we can train a linear
classifier with \emph{multiple }channels, but using only \emph{element-wise
}operations. This may be unexpected at first, since those works require
more expensive matrix inversions \cite{henriques2013beyond,galoogahi_multi-channel_2013,boddeti_correlation_2013}.

We resolve this discrepancy by pointing out that this is only possible
because we only consider a \emph{single} base sample $\mathbf{x}$.
In this case, the kernel matrix $K=XX^{T}$ is $n\times n$, regardless
of the number of features or channels. It relates the $n$ cyclic
shifts of the base sample, and can be diagonalized by the $n$ basis
of the DFT. Since $K$ is fully diagonal we can use solely element-wise
operations. However, if we consider two base samples, $K$ becomes
$2n\times2n$ and the $n$ DFT basis are no longer enough to fully
diagonalize it. This incomplete diagonalization (block-diagonalization)
requires more expensive operations to deal with, which were proposed
in those works.

With an interestingly symmetric argument, training with multiple base
samples and a single channel can be done in the primal, with only
element-wise operations (Appendix \ref{sub:appendix-MOSSE-filter}).
This follows by applying the same reasoning to the non-centered covariance
matrix $X^{T}X$, instead of $XX^{T}$. In this case we obtain the
original MOSSE filter \cite{bolme_visual_2010}.

In conclusion, for fast element-wise operations we can choose multiple
channels (in the dual, obtaining the DCF) or multiple base samples
(in the primal, obtaining the MOSSE), but not both at the same time.
This has an important impact on time-critical applications, such as
tracking. The general case \cite{henriques2013beyond} is much more
expensive and suitable mostly for offline training applications.

\section{Experiments}

\subsection{Tracking pipeline}

We implemented in Matlab two simple trackers based on the proposed
Kernelized Correlation Filter (KCF), using a Gaussian kernel, and
Dual Correlation Filter (DCF), using a linear kernel. We do not report
results for a polynomial kernel as they are virtually identical to
those for the Gaussian kernel, and require more parameters. We tested
two further variants: one that works directly on the raw pixel values,
and another that works on HOG descriptors with a cell size of 4 pixels,
in particular Felzenszwalb's variant \cite{felzenszwalb_object_2010,DollarPAMI14pyramids}.
Note that our linear DCF is equivalent to MOSSE \cite{bolme_visual_2010}
in the limiting case of a single channel (raw pixels), but it has
the advantage of also supporting multiple channels (e.g. HOG). Our
tracker requires few parameters, and we report the values that we
used, fixed for all videos, in Table \ref{tab:Parameters}.

The bulk of the functionality of the KCF is presented as Matlab code
in Algorithm \ref{alg::MATLAB-code-for}. Unlike the earlier version
of this work \cite{henriques2012circulant}, it is prepared to deal
with multiple channels, as the $3^{\textrm{rd}}$ dimension of the
input arrays. It implements 3 functions: \texttt{train} (Eq. \ref{eq:fast_krr}),
\texttt{detect} (Eq. \ref{eq:fast-detection}), and \texttt{kernel\_correlation}
(Eq. \ref{eq:gauss-multi}), which is used by the first two functions.

The pipeline for the tracker is intentionally simple, and does not
include any heuristics for failure detection or motion modeling. In
the first frame, we \texttt{train} a model with the image patch at
the initial position of the target. This patch is larger than the
target, to provide some context. For each new frame, we \texttt{detect}
over the patch at the previous position, and the target position is
updated to the one that yielded the maximum value. Finally, we \texttt{train}
a new model at the new position, and linearly interpolate the obtained
values of $\boldsymbol{\alpha}$ and $\mathbf{x}$ with the ones from
the previous frame, to provide the tracker with some memory.

\begin{algorithm}
{\small{}\medskip{}
}{\small \par}

{\small{}Inputs}{\small \par}

{\small{}\quad{}}\texttt{\small{}\textbullet \,x}{\small{}: training
image patch, $m\times n\times c$}{\small \par}

{\small{}\quad{}}\texttt{\small{}\textbullet \,y}{\small{}: regression
target, Gaussian-shaped, $m\times n$}{\small \par}

{\small{}\quad{}}\texttt{\small{}\textbullet \,z}{\small{}: test
image patch, $m\times n\times c$}{\small \par}

{\small{}Output}{\small \par}

{\small{}\quad{}}\texttt{\small{}\textbullet \,responses}{\small{}:
detection score for each location, $m\times n$\medskip{}
}{\small \par}

\begin{lstlisting}[basicstyle={\footnotesize\ttfamily},language=Matlab,moredelim={[is][\bfseries]{|}{|}},tabsize=4]
function alphaf = train(x, y, sigma, lambda)
  k = kernel_correlation(x, x, sigma);
  alphaf = fft2(y) ./ (fft2(k) + lambda);
end

function responses = detect(alphaf, x, z, sigma)
  k = kernel_correlation(z, x, sigma);
  responses = real(ifft2(alphaf .* fft2(k)));
end

function k = kernel_correlation(x1, x2, sigma)
  c = ifft2(sum(conj(fft2(x1)) .* fft2(x2), 3));
  d = x1(:)'*x1(:) + x2(:)'*x2(:) - 2 * c;
  k = exp(-1 / sigma^2 * abs(d) / |numel|(d));
end
\end{lstlisting}

\protect\caption{\textbf{:\enspace{}Matlab code, with a Gaussian kernel}.\protect \\
Multiple channels (third dimension of image patches) are supported.
It is possible to further reduce the number of FFT calls. Implementation
with GUI available at:\protect \\
\texttt{\protect\url{http://www.isr.uc.pt/~henriques/}} \label{alg::MATLAB-code-for}}
\end{algorithm}

\subsection{Evaluation}

We put our tracker to the test by using a recent benchmark that includes
50 video sequences \cite{y_wu_online_2013} (see Fig. \ref{fig:Qualitative-results-for}).
This dataset collects many videos used in previous works, so we avoid
the danger of overfitting to a small subset.

For the performance criteria, we did not choose average location error
or other measures that are averaged over frames, since they impose
an arbitrary penalty on lost trackers that depends on chance factors
(i.e., the position where the track was lost), making them not comparable.
A similar alternative is bounding box overlap, which has the disadvantage
of heavily penalizing trackers that do not track across scale, even
if the target position is otherwise tracked perfectly.

An increasingly popular alternative, which we chose for our evaluation,
is the precision curve \cite{y_wu_online_2013,babenko_robust_2011,henriques2012circulant}.
A frame may be considered correctly tracked if the predicted target
center is within a distance threshold of ground truth. Precision curves
simply show the percentage of correctly tracked frames for a range
of distance thresholds. Notice that by plotting the precision for
all thresholds, no parameters are required. This makes the curves
unambiguous and easy to interpret. A higher precision at low thresholds
means the tracker is more accurate, while a lost target will prevent
it from achieving perfect precision for a very large threshold range.
When a representative precision score is needed, the chosen threshold
is 20 pixels, as done in previous works \cite{y_wu_online_2013,babenko_robust_2011,henriques2012circulant}.

\begin{figure}
\begin{centering}
\includegraphics[width=0.99\columnwidth]{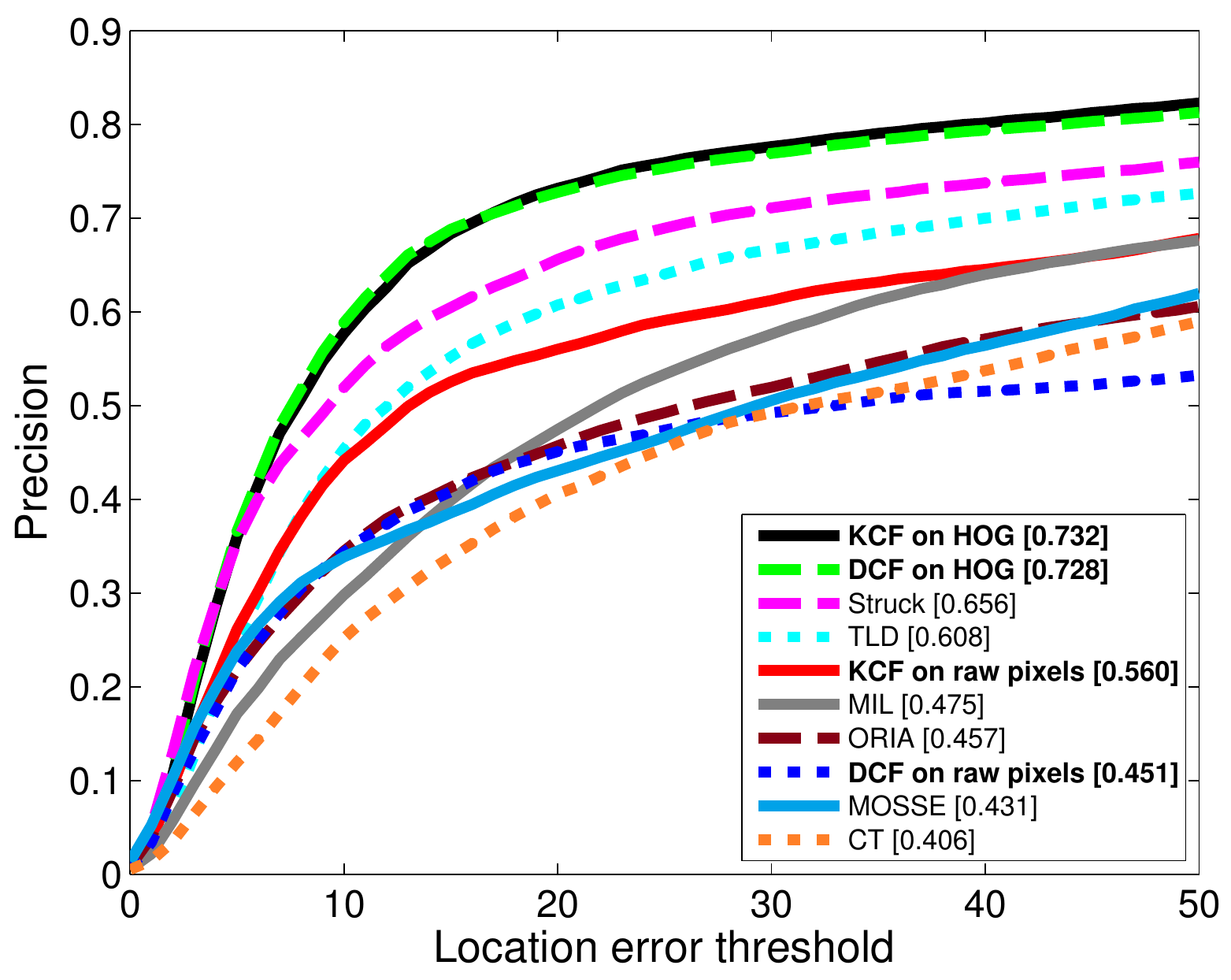}
\par\end{centering}

\protect\caption{Precision plot for all 50 sequences. The proposed trackers (bold)
outperform state-of-the-art systems, such as TLD and Struck, which
are more complicated to implement and much slower (see Table \ref{tab:Summary-of-experimental}).
Best viewed in color.\label{fig:Precision-plot-all}}
\end{figure}

\pagebreak{}

\subsection{Experiments on the full dataset}

We start by summarizing the results over all videos in Table \ref{tab:Summary-of-experimental}
and Fig. \ref{fig:Precision-plot-all}. For comparison, we also report
results for several other systems \cite{hare_struck:_2011,kalal_2012,bolme_visual_2010,babenko_robust_2011,wu_online_2012,zhang_real-time_2012},
including some of the most resilient trackers available -- namely,
Struck and TLD. Unlike our simplistic implementation (Algorithm \ref{alg::MATLAB-code-for}),
these trackers contain numerous engineering improvements. Struck operates
on many different kinds of features and a growing pool of support
vectors. TLD is specifically geared towards re-detection, using a
set of structural rules with many parameters.

Despite this asymmetry, our Kernelized Correlation Filter (KCF) can
already reach competitive performance by operating on raw pixels alone,
as can be seen in Fig. \ref{fig:Precision-plot-all}. In this setting,
the rich implicit features induced by the Gaussian kernel yield a
distinct advantage over the proposed Dual Correlation Filter (DCF).

We remark that the DCF with single-channel features (raw pixels) is
theoretically equivalent to a MOSSE filter \cite{bolme_visual_2010}.
For a direct comparison, we include the results for the authors' MOSSE
tracker \cite{bolme_visual_2010} in Fig. \ref{fig:Precision-plot-all}.
The performance of both is very close, showing that any particular
differences in their implementations do not seem to matter much. However,
the kernelized algorithm we propose (KCF) does yield a noticeable
increase in performance.

Replacing pixels with HOG features allows the KCF and DCF to surpass
even TLD and Struck, by a relatively large margin (Fig. \ref{fig:Precision-plot-all}).
This suggests that the most crucial factor for high performance, compared
to other trackers that use similar features, is the efficient incorporation
of thousands of negative samples from the target's environment, which
they do with very little overhead.

\begin{table}
\begin{tabular}{|>{\centering}m{1.3cm}|>{\centering}m{1.8cm}|>{\centering}m{1.2cm}|>{\centering}m{1.3cm}|>{\centering}m{0.8cm}|}
\cline{2-5} 
\multicolumn{1}{>{\centering}m{1.3cm}|}{} & Algorithm & Feature & Mean precision\\
(20 px) & Mean FPS\tabularnewline
\hline 
\multirow{4}{1.3cm}{Proposed} & KCF & \multirow{2}{1.2cm}{HOG} & \textbf{73.2\%} & 172\tabularnewline
\cline{2-2} \cline{4-5} 
 & DCF &  & \textbf{72.8\%} & \textbf{292}\tabularnewline
\cline{2-5} 
 & KCF & \multirow{2}{1.2cm}{Raw pixels} & 56.0\% & 154\tabularnewline
\cline{2-2} \cline{4-5} 
 & DCF &  & 45.1\% & 278\tabularnewline
\hline 
\hline 
\multirow{6}{1.3cm}{Other algorithms} & \multicolumn{2}{c|}{Struck \cite{hare_struck:_2011}} & 65.6\% & ~20\tabularnewline
\cline{2-5} 
 & \multicolumn{2}{c|}{TLD \cite{kalal_2012}} & 60.8\% & ~28\tabularnewline
\cline{2-5} 
 & \multicolumn{2}{c|}{MOSSE \cite{bolme_visual_2010}} & 43.1\% & \textbf{615}\tabularnewline
\cline{2-5} 
 & \multicolumn{2}{c|}{MIL \cite{babenko_robust_2011}} & 47.5\% & ~38\tabularnewline
\cline{2-5} 
 & \multicolumn{2}{c|}{ORIA \cite{wu_online_2012}} & 45.7\% & ~~9\tabularnewline
\cline{2-5} 
 & \multicolumn{2}{c|}{CT \cite{zhang_real-time_2012}} & 40.6\% & ~64\tabularnewline
\hline 
\end{tabular}

\protect\caption{Summary of experimental results on the 50 videos dataset. The reported
quantities are averaged over all videos. Reported speeds include feature
computation (e.g. HOG).\label{tab:Summary-of-experimental}}
\end{table}

\medskip{}

\noindent \textbf{Timing.}\enspace{}As mentioned earlier, the overall
complexity of our closed-form solutions is $\mathcal{O}\left(n\log n\right)$,
resulting in its high speed (Table \ref{tab:Summary-of-experimental}).
The speed of the tracker is directly related to the size of the tracked
region. This is an important factor when comparing trackers based
on correlation filters. MOSSE \cite{bolme_visual_2010} tracks a region
that has the same support as the target object, while our implementation
tracks a region that is 2.5 times larger (116x170 on average). Reducing
the tracked region would allow us to approach their FPS of 615 (Table
\ref{tab:Summary-of-experimental}), but we found that it hurts performance,
especially for the kernel variants. Another interesting observation
from Table \ref{tab:Summary-of-experimental} is that operating on
31 HOG features per spatial cell can be slightly faster than operating
on raw pixels, even though we take the overhead of computing HOG features
into account. Since each 4x4 pixels cell is represented by a single
HOG descriptor, the smaller-sized DFTs counterbalance the cost of
iterating over feature channels. Taking advantage of all 4 cores of
a desktop computer, KCF/DCF take less than 2 minutes to process all
50 videos ($\sim$29,000 frames).

\begin{figure*}
\begin{centering}
\hfill{}\includegraphics[width=0.75\columnwidth]{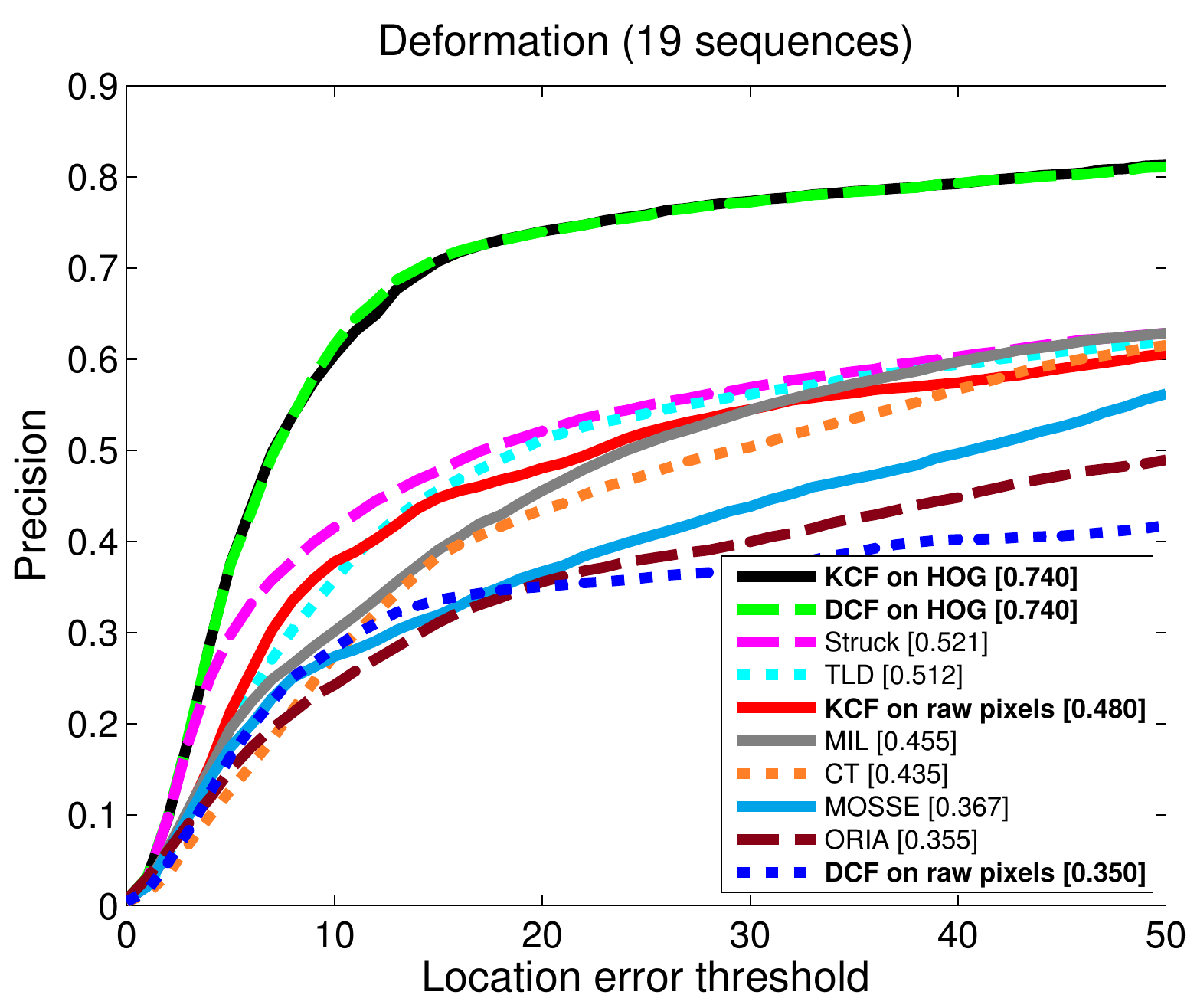}\hfill{}\hfill{}\includegraphics[width=0.75\columnwidth]{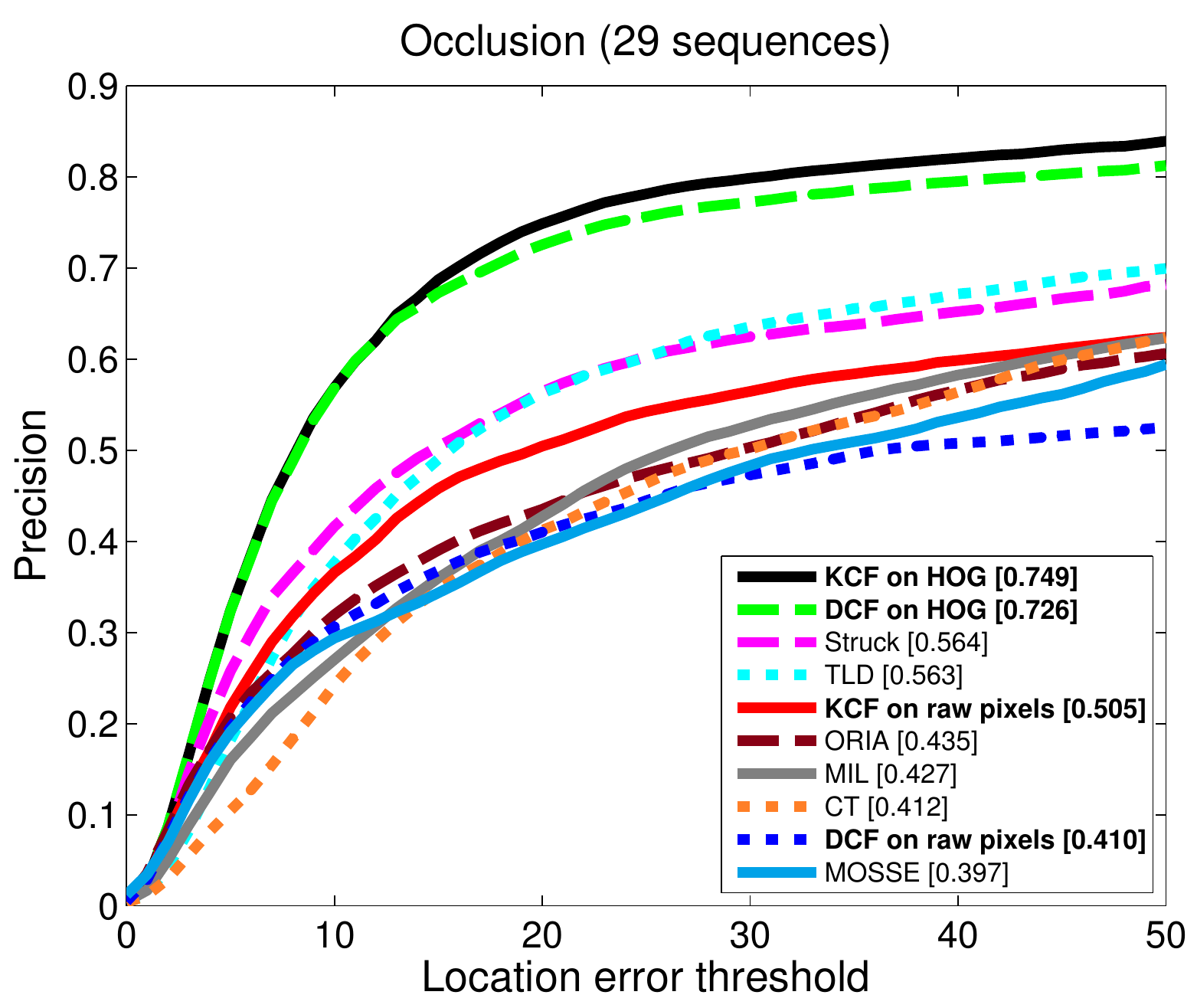}\hfill{}
\par\end{centering}

\medskip{}

\begin{centering}
\hfill{}\includegraphics[width=0.75\columnwidth]{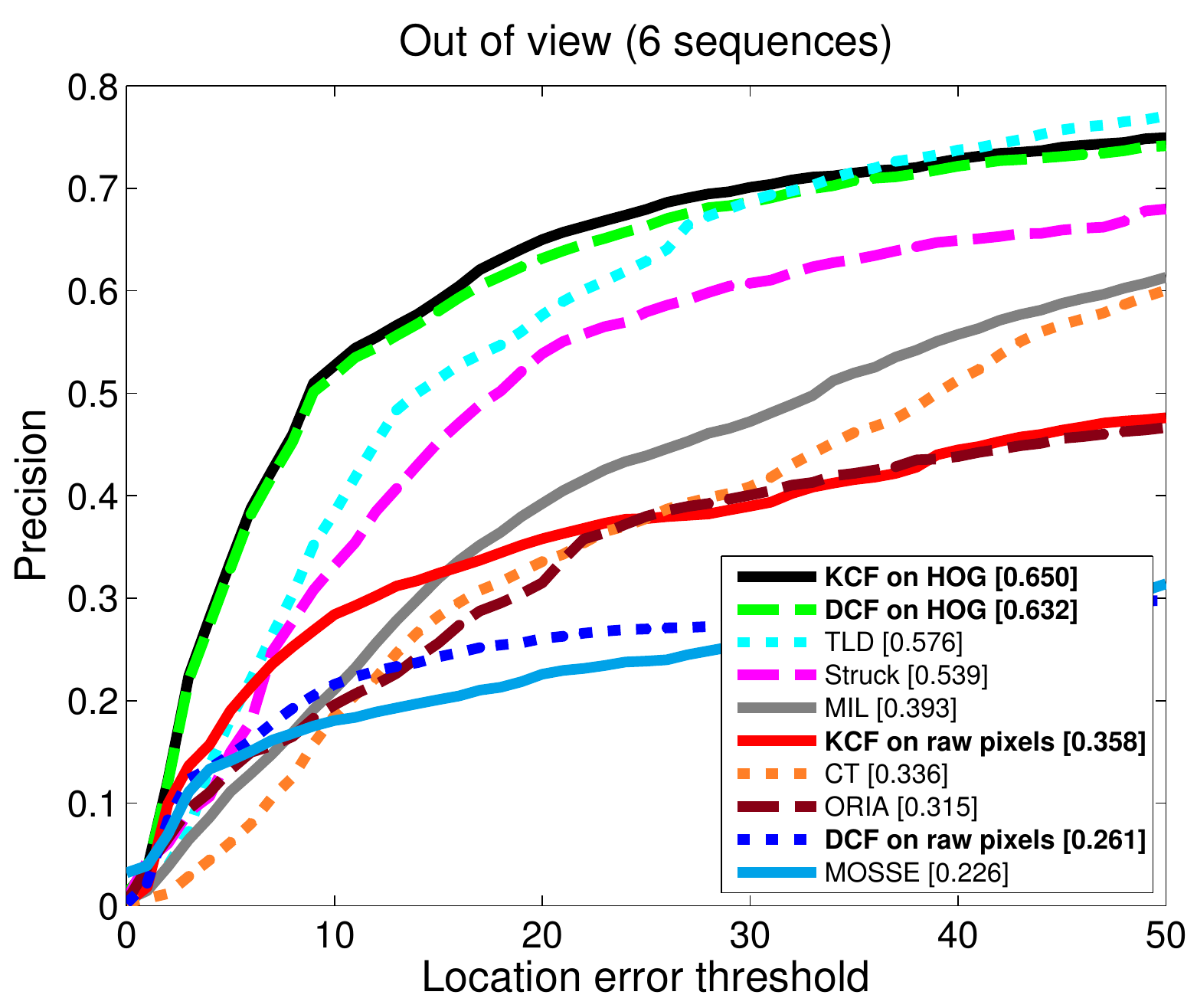}\hfill{}\hfill{}\includegraphics[width=0.75\columnwidth]{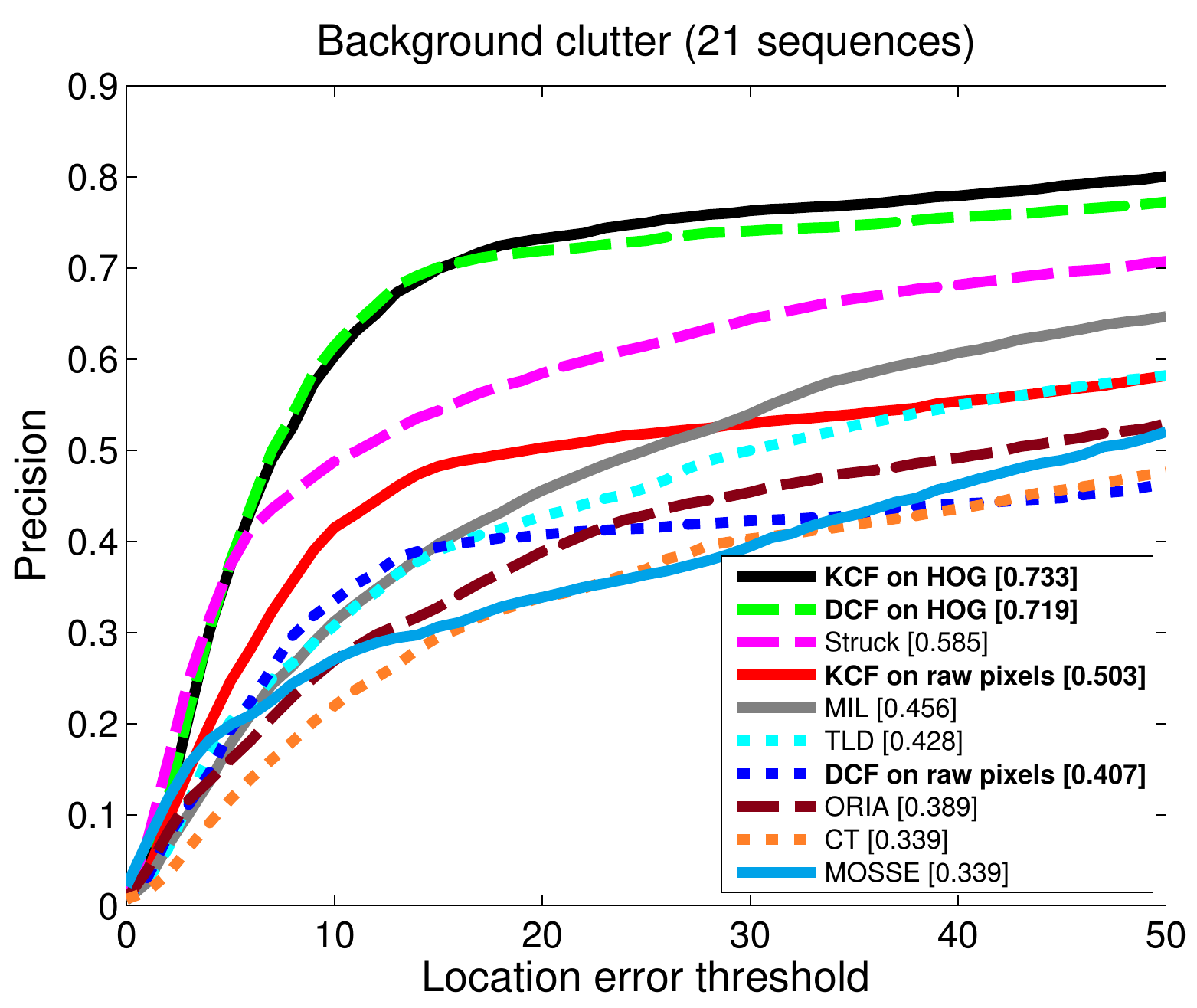}\hfill{}
\par\end{centering}

\protect\caption{Precision plot for sequences with attributes: occlusion, non-rigid
deformation, out-of-view target, and background clutter. The HOG variants
of the proposed trackers (bold) are the most resilient to all of these
nuisances. Best viewed in color.\label{fig:Precision-attributes}}
\end{figure*}

\subsection{Experiments with sequence attributes}

The videos in the benchmark dataset \cite{y_wu_online_2013} are annotated
with attributes, which describe the challenges that a tracker will
face in each sequence -- e.g., illumination changes or occlusions.
These attributes are useful for diagnosing and characterizing the
behavior of trackers in such a large dataset, without having to analyze
each individual video. We report results for 4 attributes in Figure
\ref{fig:Precision-attributes}: non-rigid deformations, occlusions,
out-of-view target, and background clutter.

The robustness of the HOG variants of our tracker regarding non-rigid
deformations and occlusions is not surprising, since these features
are known to be highly discriminative \cite{felzenszwalb_object_2010}.
However, the KCF on raw pixels alone still fares almost as well as
Struck and TLD, with the kernel making up for the features' shortcomings.

One challenge for the system we implemented is an out-of-view target,
due to the lack of a failure recovery mechanism. TLD performs better
than most other trackers in this case, which illustrates its focus
on re-detection and failure recovery. Such engineering improvements
could probably benefit our trackers, but the fact that KCF/DCF can
still outperform TLD shows that they are not a decisive factor.

Background clutter severely affects almost all trackers, except for
the proposed ones, and to a lesser degree, Struck. For our tracker
variants, this is explained by the implicit inclusion of thousands
of negative samples around the tracked object. Since in this case
even the raw pixel variants of our tracker have a performance very
close to optimal, while TLD, CT, ORIA and MIL show degraded performance,
we conjecture that this is caused by their undersampling of negatives.

\begin{figure*}
\begin{centering}
\includegraphics[width=0.31\textwidth]{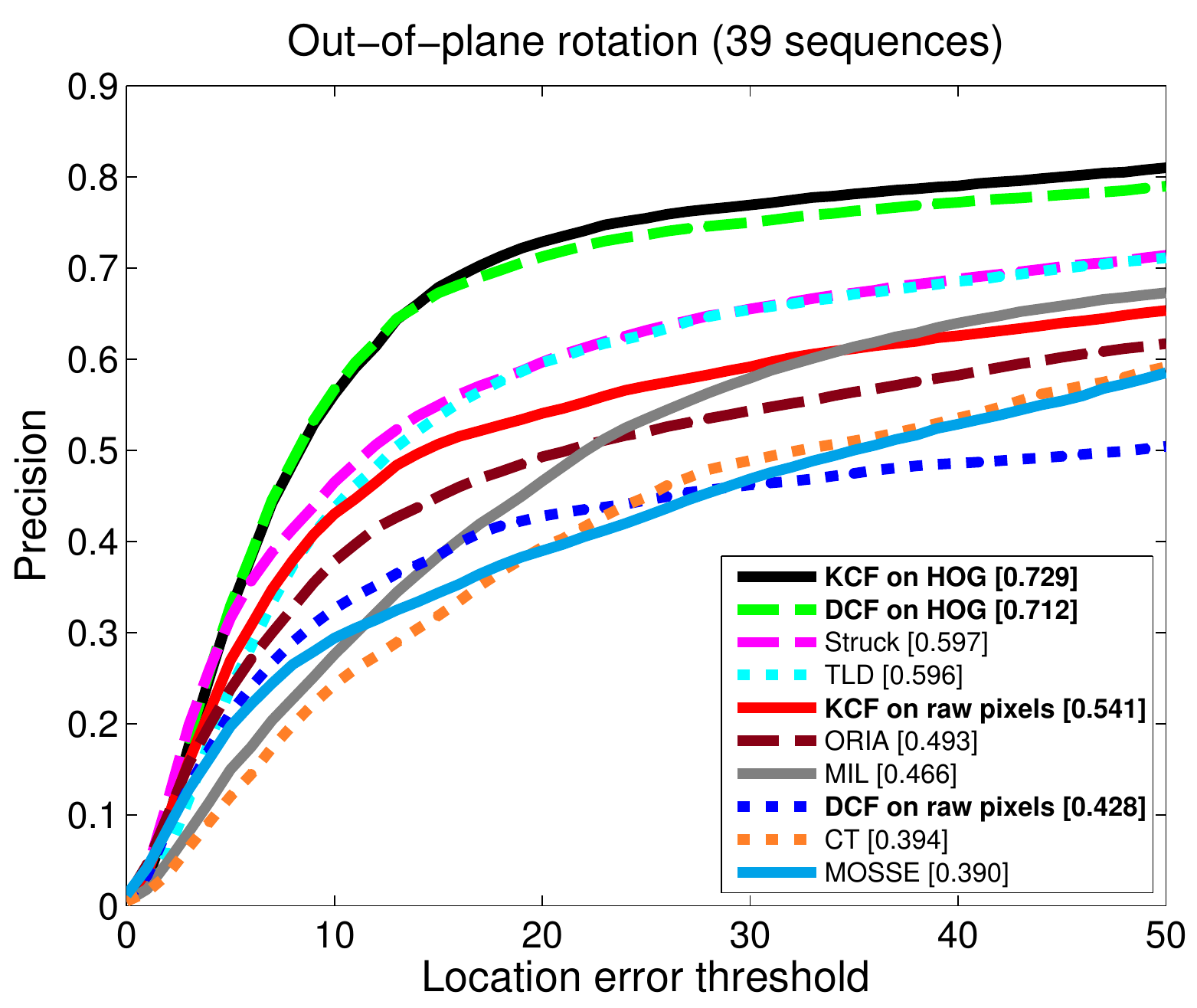}\hfill{}\includegraphics[width=0.31\textwidth]{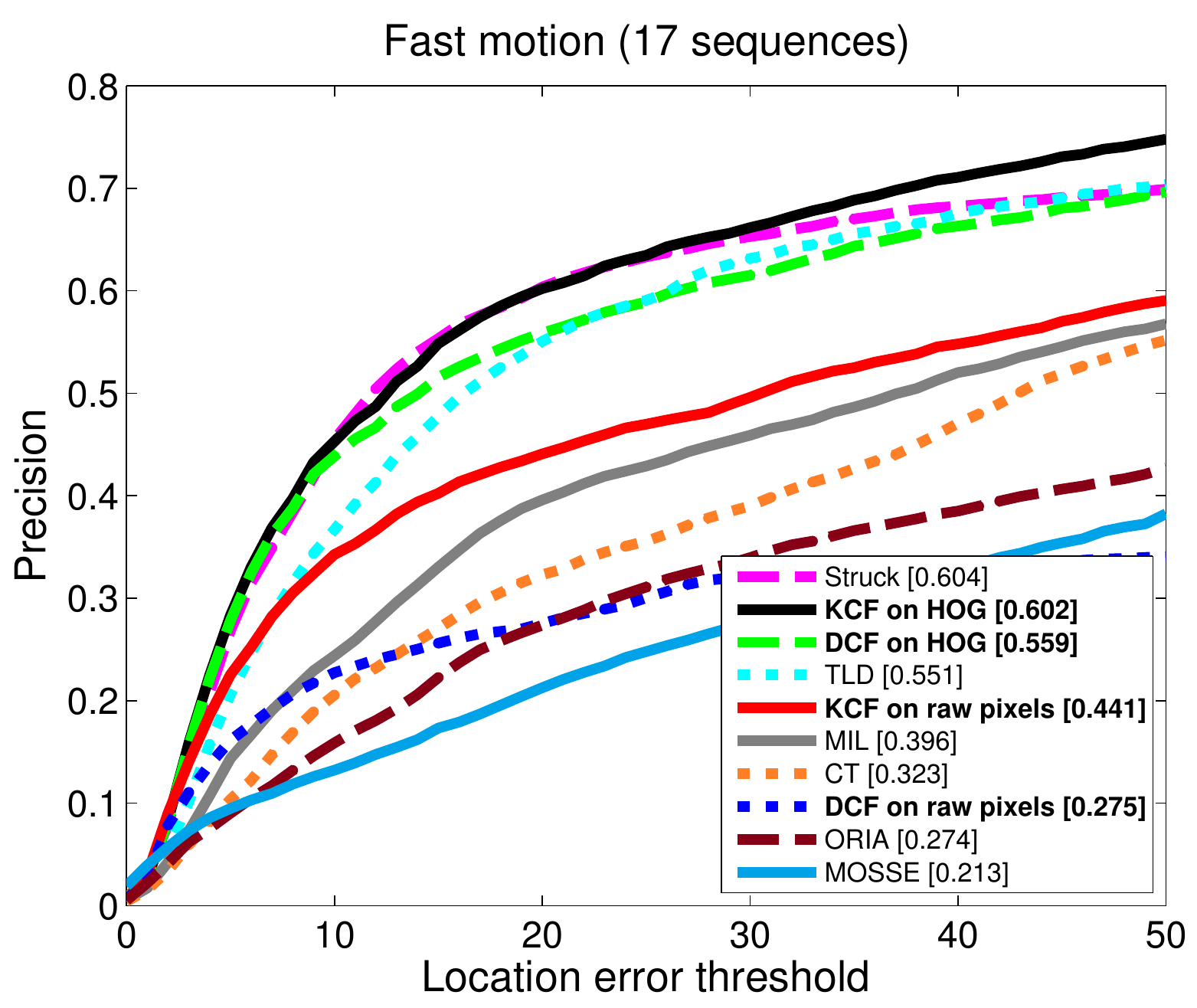}\hfill{}\includegraphics[width=0.31\textwidth]{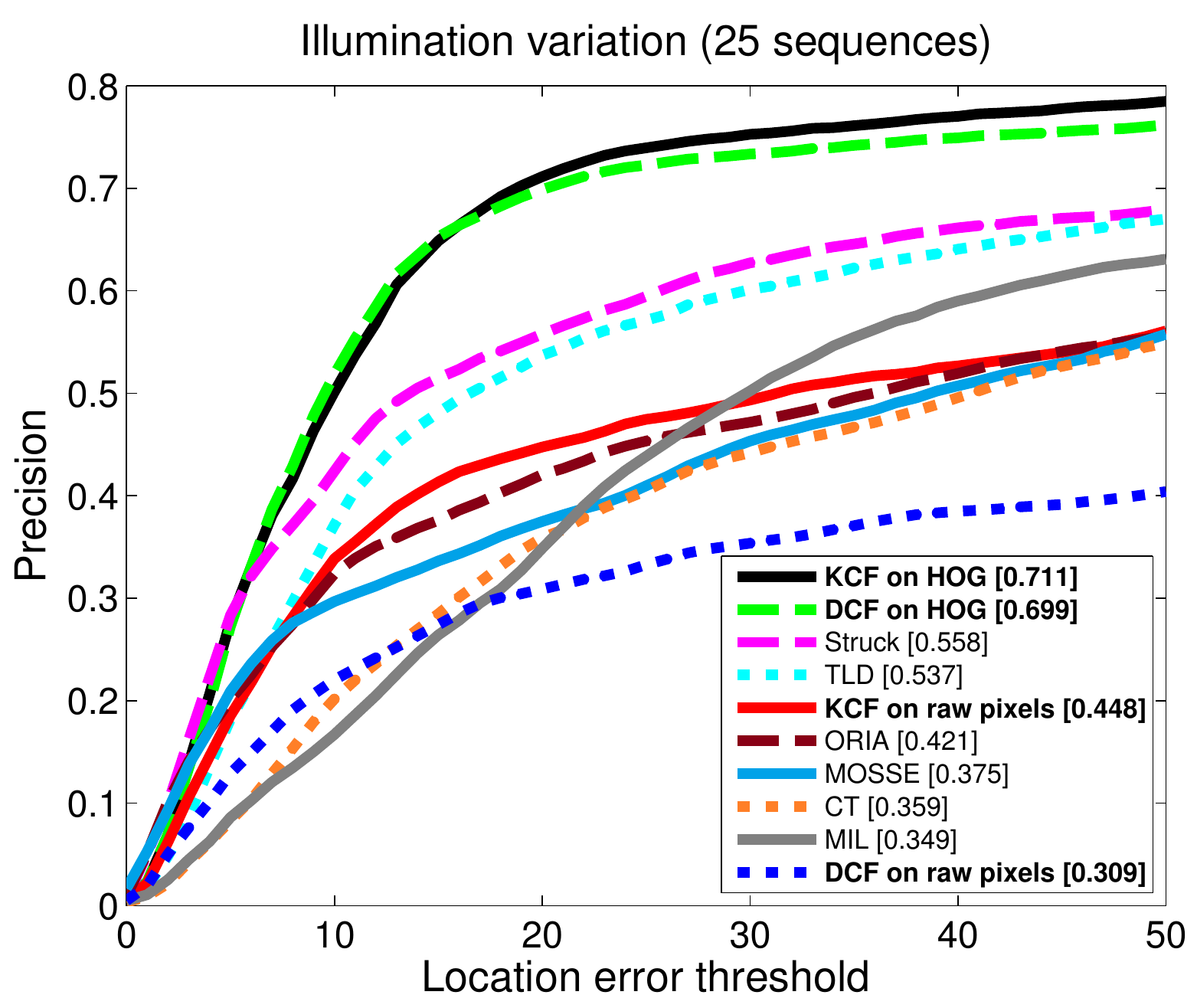}\medskip{}

\par\end{centering}

\begin{centering}
\includegraphics[width=0.31\textwidth]{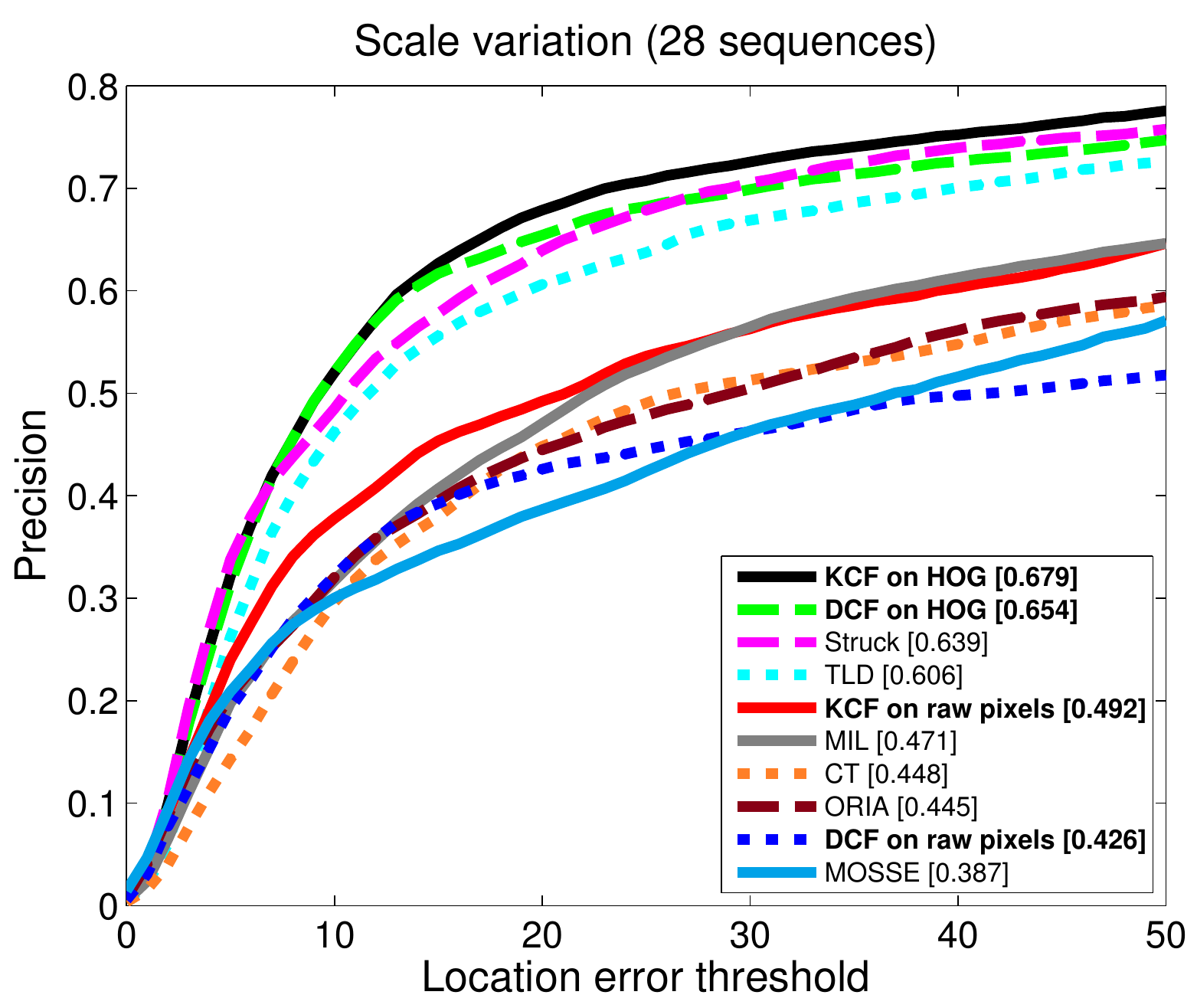}\hfill{}\includegraphics[width=0.31\textwidth]{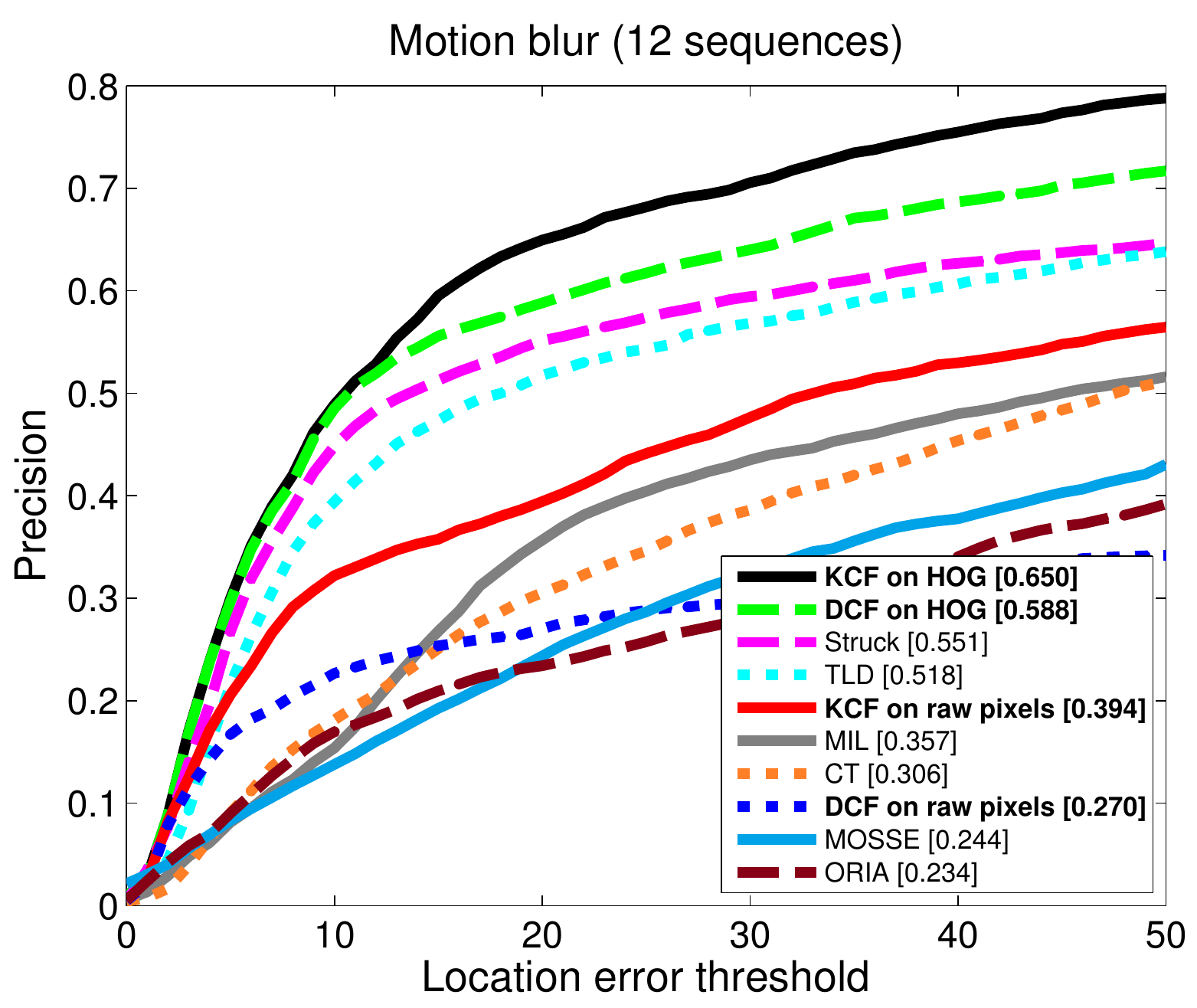}\hfill{}\includegraphics[width=0.31\textwidth]{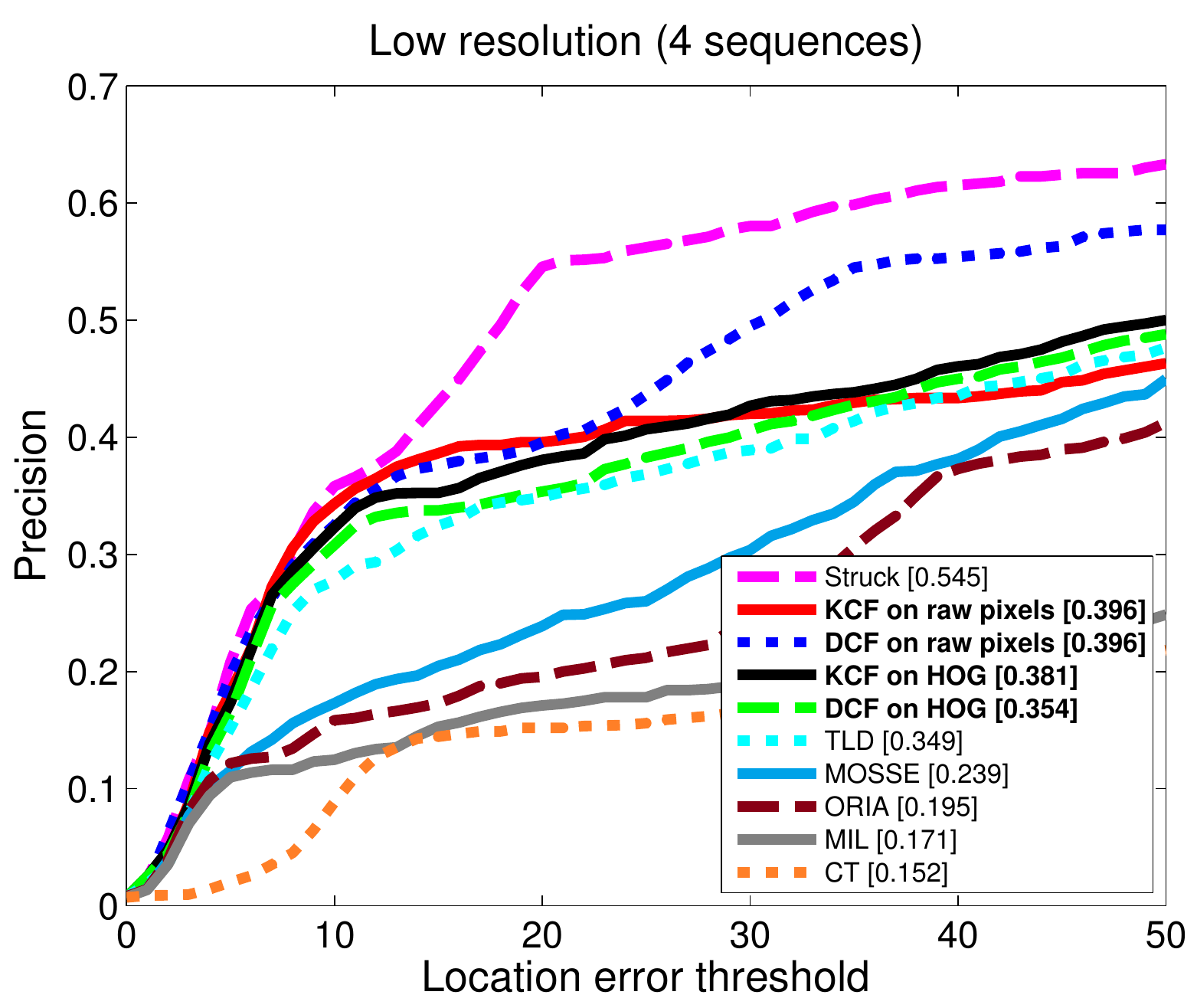}
\par\end{centering}

\medskip{}

\protect\caption{Precision plots for 6 attributes of the dataset. Best viewed in color.
In-plane rotation was left out due to space constraints. Its results
are virtually identical to those for out-of-plane rotation (above),
since they share almost the same set of sequences.\label{fig:more-attributes}}
\end{figure*}

We also report results for other attributes in Fig. \ref{fig:more-attributes}.
Generally, the proposed trackers are the most robust to 6 of the 7
challenges, except for low resolution, which affects equally all trackers
but Struck.

\section{Conclusions and future work}

In this work, we demonstrated that it is possible to analytically
model natural image translations, showing that under some conditions
the resulting data and kernel matrices become circulant. Their diagonalization
by the DFT provides a general blueprint for creating fast algorithms
that deal with translations. We have applied this blueprint to linear
and kernel ridge regression, obtaining state-of-the-art trackers that
run at hundreds of FPS and can be implemented with only a few lines
of code. Extensions of our basic approach seem likely to be useful
in other problems. Since the first version of this work, circulant
data has been exploited successfully for other algorithms in detection
\cite{henriques2013beyond} and video event retrieval \cite{revaud_event_2013}.
An interesting direction for further work is to relax the assumption
of periodic boundaries, which may improve performance. Many useful
algorithms may also be obtained from the study of other objective
functions with circulant data, including classical filters such as
SDF or MACE \cite{patnaik_fast_2009,jeong2006kernel}, and more robust
loss functions than the squared loss. We also hope to generalize this
framework to other operators, such as affine transformations or non-rigid
deformations.

\section*{Acknowledgment}

\noindent The authors acknowledge support by the FCT project PTDC/EEA-CRO/122812/2010,
grants SFRH/BD75459/2010, SFRH/BD74152/2010, and SFRH/BPD/90200/2012.

\appendices{}

\section{~}

\subsection{Implementation details\label{sub:Implementation-details}}

As is standard with correlation filters, the input patches (either
raw pixels or extracted feature channels) are weighted by a cosine
window, which smoothly removes discontinuities at the image boundaries
caused by the cyclic assumption \cite{bolme_visual_2010,gonzalez_digital_2008}.
The tracked region has 2.5 times the size of the target, to provide
some context and additional negative samples.

Recall that the training samples consist of shifts of a base sample,
so we must specify a regression target for each one in $\mathbf{y}$.
The regression targets $\mathbf{y}$ simply follow a Gaussian function,
which takes a value of 1 for a centered target, and smoothly decays
to 0 for any other shifts, according to the spatial bandwidth $s$.
Gaussian targets are smoother than binary labels, and have the benefit
of reducing ringing artifacts in the Fourier domain \cite{gonzalez_digital_2008}.

\begin{figure}
\begin{centering}
\hfill{}\includegraphics[width=0.35\columnwidth]{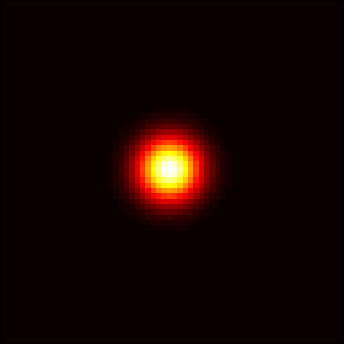}\hfill{}\hfill{}\includegraphics[width=0.35\columnwidth]{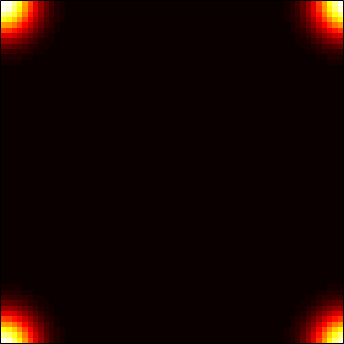}\hfill{}
\par\end{centering}

\begin{centering}
{\small{}\hfill{}(a)\hfill{}\hfill{}(b)\hfill{}}
\par\end{centering}{\small \par}

\protect\caption{Regression targets $\mathbf{y}$, following a Gaussian function with
spatial bandwidth $s$ (white indicates a value of 1, black a value
of 0). (a) Placing the peak in the middle will unnecessarily cause
the detection output to be shifted by half a window (discussed in
Section \ref{sub:Implementation-details}). (b) Placing the peak at
the top-left element (and wrapping around) correctly centers the detection
output. \label{fig:Regression-targets}}
\end{figure}

A subtle issue is determining which element of $\mathbf{y}$ is the
regression target for the centered sample, on which we will center
the Gaussian function. Although intuitively it may seem to be the
middle of the output plane (Fig. \ref{fig:Regression-targets}-a),
it turns out that the correct choice is the top-left element (Fig.
\ref{fig:Regression-targets}-b). The explanation is that, after computing
a cross-correlation between two images in the Fourier domain and converting
back to the spatial domain, it is the top-left element of the result
that corresponds to a shift of zero \cite{gonzalez_digital_2008}.
Of course, since we always deal with cyclic signals, the peak of the
Gaussian function must wrap around from the top-left corner to the
other corners, as can be seen in Fig. \ref{fig:Regression-targets}-b.
Placing the Gaussian peak in the middle of the regression target is
common in some filter implementations, and leads the correlation output
to be unnecessarily shifted by half a window, which must be corrected
post-hoc%
\footnote{This is usually done by switching the quadrants of the output, e.g.
with the Matlab built-in function \texttt{fftshift}. It has the same
effect as shifting Fig. \ref{fig:Regression-targets}-b to look like
Fig. \ref{fig:Regression-targets}-a.%
}.

Another common source of error is the fact that most implementations
of the Fast Fourier Transform%
\footnote{For example Matlab, NumPy, Octave and the FFTW library.%
} do not compute the unitary DFT. This means that the L2 norm of the
signals is not preserved, unless the output is corrected by a constant
factor. With some abuse of notation, we can say that the unitary DFT
may be computed as

\[
\mathcal{F}_{\mathcal{U}}(x)=\mathtt{fft2(x)~/~sqrt(m*n)},
\]
where the input $x$ has size $m\times n$, and similarly for the
inverse DFT,

\[
\mathcal{F}_{\mathcal{U}}^{-1}(x)=\mathtt{ifft2(x)~*~sqrt(m*n)}.
\]

\begin{table}
\begin{tabular}{|>{\centering}m{3.4cm}|>{\centering}m{2.3cm}|>{\centering}m{1.8cm}|}
\hline 
KCF/DCF parameters & With raw pixels & With HOG\tabularnewline
\hline 
Feature bandwidth $\sigma$ & 0.2 & 0.5\tabularnewline
\hline 
Adaptation rate & 0.075 & 0.02\tabularnewline
\hline 
Spatial bandwidth $s$ & \multicolumn{2}{c|}{$\sqrt{mn}/10$}\tabularnewline
\hline 
Regularization $\lambda$ & \multicolumn{2}{c|}{$10^{-4}$}\tabularnewline
\hline 
\end{tabular}

\protect\caption{Parameters used in all experiments. In this table, $n$ and $m$ refer
to the width and height of the target, measured in pixels or HOG cells.\label{tab:Parameters}}
\end{table}

\subsection{Proof of Theorem \ref{thm:circulant-K}\label{sub:Proof-of-Theorem}}

Under the theorem's assumption that $\kappa(\mathbf{x},\mathbf{x}')=\kappa(M\mathbf{x},M\mathbf{x}')$,
for any permutation matrix $M$, then

\begin{align*}
K_{ij} & =\kappa(P^{i}\mathbf{x},\, P^{j}\mathbf{x})\\
 & =\kappa(P^{-i}P^{i}\mathbf{x},\, P^{-i}P^{j}\mathbf{x}).
\end{align*}

Using known properties of permutation matrices, this reduces to

\begin{equation}
K_{ij}=\kappa(\mathbf{x},\, P^{j-i}\mathbf{x}).\label{eq:theorem_proof_K1}
\end{equation}

By the cyclic nature of $P$, it repeats every $n$th power, i.e.
$P^{n}=P^{0}$. As such, Eq. \ref{eq:theorem_proof_K1} is equivalent
to

\begin{equation}
K_{ij}=\kappa(\mathbf{x},\, P^{(j-i)\textrm{ mod }n}\,\mathbf{x}),\label{eq:theorem_proof_K2}
\end{equation}
where mod is the modulus operation (remainder of division by $n$).

We now use the fact the elements of a circulant matrix $X=C(\mathbf{x})$
(Eq. \ref{eq:X}) satisfy

\[
X_{ij}=x_{\left((j-i)\textrm{ mod }n\right)+1},
\]
that is, a matrix is circulant if its elements only depend on $(j-i)\textrm{ mod }n$.
It is easy to check that this condition is satisfied by Eq. \ref{eq:X},
and in fact it is often used as the definition of a circulant matrix
\cite{gray_toeplitz_2006}.

Because $K_{ij}$ also depends on $(j-i)\textrm{ mod }n$, we must
conclude that $K$ is circulant as well, finishing our proof.

\subsection{Kernel Ridge Regression with Circulant data\label{sub:appendix-Kernel-Ridge-Regression}}

This section shows a more detailed derivation of Eq. \ref{eq:fast_krr}.
We start by replacing  $K=C(\mathbf{k}^{\mathbf{xx}})$ in the formula
for Kernel Ridge Regression, Eq. \ref{eq:krr}, and diagonalizing
it

\begin{align*}
\boldsymbol{\alpha} & =\left(C(\mathbf{k}^{\mathbf{xx}})+\lambda I\right)^{-1}\mathbf{y}\\
 & =\left(F\textrm{diag}\left(\hat{\mathbf{k}}^{\mathbf{xx}}\right)F^{H}+\lambda I\right)^{-1}\mathbf{y}.
\end{align*}

By simple linear algebra, and the unitarity of $F$ ($FF^{H}=I$),

\begin{align*}
\boldsymbol{\alpha} & =\left(F\textrm{diag}\left(\hat{\mathbf{k}}^{\mathbf{xx}}\right)F^{H}+\lambda FIF^{H}\right)^{-1}\mathbf{y}\\
 & =F\textrm{diag}\left(\hat{\mathbf{k}}^{\mathbf{xx}}+\lambda\right)^{-1}F^{H}\mathbf{y},
\end{align*}
which is equivalent to

\[
F^{H}\boldsymbol{\alpha}=\textrm{diag}\left(\hat{\mathbf{k}}^{\mathbf{xx}}+\lambda\right)^{-1}F^{H}\mathbf{y}.
\]
Since for any vector $F\mathbf{z}=\hat{\mathbf{z}}$, we have

\[
\hat{\boldsymbol{\alpha}}^{*}=\textrm{diag}\left(\frac{1}{\hat{\mathbf{k}}^{\mathbf{xx}}+\lambda}\right)\hat{\mathbf{y}}^{*}.
\]

Finally, because the product of a diagonal matrix and a vector is
simply their element-wise product,

\[
\hat{\boldsymbol{\alpha}}^{*}=\frac{\hat{\mathbf{y}}^{*}}{\hat{\mathbf{k}}^{\mathbf{xx}}+\lambda}.
\]

\subsection{Derivation of fast detection formula\label{sub:appendix-detection}}

To diagonalize Eq. \ref{eq:detection}, we use the same properties
as in the previous section. We have

\begin{align*}
\mathbf{f}(\mathbf{z}) & =\left(C\left(\mathbf{k}^{\mathbf{xz}}\right)\right)^{T}\boldsymbol{\alpha}\\
 & =\left(F\textrm{diag}\left(\hat{\mathbf{k}}^{\mathbf{xz}}\right)F^{H}\right)^{T}\boldsymbol{\alpha}\\
 & =F^{H}\textrm{diag}\left(\hat{\mathbf{k}}^{\mathbf{xz}}\right)F\boldsymbol{\alpha}
\end{align*}
which is equivalent to

\[
F\mathbf{f}(\mathbf{z})=\textrm{diag}\left(\hat{\mathbf{k}}^{\mathbf{xz}}\right)F\boldsymbol{\alpha}.
\]

Replicating the same final steps from the previous section,

\[
\hat{\mathbf{f}}(\mathbf{z})=\hat{\mathbf{k}}^{\mathbf{xz}}\odot\hat{\boldsymbol{\alpha}}.
\]

\subsection{Linear Ridge Regression with Circulant data\label{sub:appendix-linear}}

This is a more detailed version of the steps from Section \ref{sub:Putting-it-all}.
It is very similar to the kernel case. We begin by replacing  Eq.
\ref{eq:non-centered-cov} in the formula for Ridge Regression, Eq.
\ref{eq:linear_ridge_regression}.

\[
\mathbf{w}=\left(F\textrm{diag}\left(\hat{\mathbf{x}}^{*}\odot\hat{\mathbf{x}}\right)F^{H}+\lambda I\right)^{-1}X^{H}\mathbf{y}
\]

By simple algebra, and the unitarity of $F$, we have

\begin{align*}
\mathbf{w} & =\left(F\textrm{diag}\left(\hat{\mathbf{x}}^{*}\odot\hat{\mathbf{x}}\right)F^{H}+\lambda F^{H}IF\right)^{-1}X^{H}\mathbf{y}\\
 & =\left(F\textrm{diag}\left(\hat{\mathbf{x}}^{*}\odot\hat{\mathbf{x}}+\lambda\right)^{-1}F^{H}\right)X^{H}\mathbf{y}\\
 & =F\textrm{diag}\left(\hat{\mathbf{x}}^{*}\odot\hat{\mathbf{x}}+\lambda\right)^{-1}F^{H}F\textrm{diag}\left(\hat{\mathbf{x}}\right)F^{H}\mathbf{y}\\
 & =F\textrm{diag}\left(\frac{\hat{\mathbf{x}}}{\hat{\mathbf{x}}^{*}\odot\hat{\mathbf{x}}+\lambda}\right)F^{H}\mathbf{y}.
\end{align*}

Then, this is equivalent to

\[
F\mathbf{w}=\textrm{diag}\left(\frac{\hat{\mathbf{x}}^{*}}{\hat{\mathbf{x}}^{*}\odot\hat{\mathbf{x}}+\lambda}\right)F\mathbf{y},
\]
and since for any vector $F\mathbf{z}=\hat{\mathbf{z}}$,

\[
\hat{\mathbf{w}}=\textrm{diag}\left(\frac{\hat{\mathbf{x}}^{*}}{\hat{\mathbf{x}}^{*}\odot\hat{\mathbf{x}}+\lambda}\right)\hat{\mathbf{y}}.
\]

We may go one step further, since the product of a diagonal matrix
and a vector is just their element-wise product.

\[
\hat{\mathbf{w}}=\frac{\hat{\mathbf{x}}^{*}\odot\hat{\mathbf{y}}}{\hat{\mathbf{x}}^{*}\odot\hat{\mathbf{x}}+\lambda}
\]

\subsection{MOSSE filter\label{sub:appendix-MOSSE-filter}}

The only difference between Eq. \ref{eq:mosse} and the MOSSE filter
\cite{bolme_visual_2010} is that the latter minimizes the error over
(cyclic shifts of) multiple base samples $\mathbf{x}_{i}$, while
Eq. \ref{eq:mosse} is defined for a single base sample $\mathbf{x}$.
This was done for clarity of presentation, and the general case is
easily derived. Note also that MOSSE does not support multiple channels,
which we do through our dual formulation.

The cyclic shifts of each base sample $\mathbf{x}_{i}$ can be expressed
in a circulant matrix $X_{i}$. Then, replacing the data matrix $X'=\left[\begin{array}{c}
X_{1}\\
X_{2}\\
\vdots
\end{array}\right]$ in Eq. \ref{eq:linear_ridge_regression} results in

\[
\mathbf{w}=\sum_{j}\left(\sum_{i}X_{i}^{H}X_{i}+\lambda I\right)^{-1}X_{j}^{H}\mathbf{y},
\]
by direct application of the rule for products of block matrices.
Factoring the bracketed expression,

\begin{equation}
\mathbf{w}=\left(\sum_{i}X_{i}^{H}X_{i}+\lambda I\right)^{-1}\left(\sum_{i}X_{i}^{H}\right)\mathbf{y}.\label{eq:appendix_mosse}
\end{equation}

Eq. \ref{eq:appendix_mosse} looks exactly like Eq. \ref{eq:linear_ridge_regression},
except for the sums. It is then trivial to follow the same steps as
in Section \ref{sub:Putting-it-all} to diagonalize it, and obtain
the filter equation

\[
\hat{\mathbf{w}}=\frac{\sum_{i}\hat{\mathbf{x}}_{i}^{*}\odot\hat{\mathbf{y}}}{\sum_{i}\hat{\mathbf{x}}_{i}^{*}\odot\hat{\mathbf{x}}_{i}+\lambda}.
\]

\vspace{-0.5cm}

\begin{IEEEbiography}[{\includegraphics[clip,width=1in,height=1.25in,keepaspectratio]{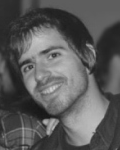}}]{João F. Henriques}
 received his M.Sc. degree in Electrical Engineering from the University
of Coimbra, Portugal in 2009. He is currently a Ph.D. student at the
Institute of Systems and Robotics, University of Coimbra. His current
research interests include Fourier analysis and machine learning algorithms
in general, with computer vision applications in detection and tracking.
\end{IEEEbiography}

\begin{IEEEbiographynophoto}{Rui Caseiro}
 received the B.Sc. degree in electrical engineering (specialization
in automation) from the University of Coimbra, Coimbra, Portugal,
in 2005. Since 2007, he has been involved in several research projects,
which include the European project \textquoteleft \textquoteleft Perception
on Purpose\textquoteright \textquoteright{} and the National project
\textquoteleft \textquoteleft Brisa-ITraffic\textquoteright \textquoteright .
He is currently a Ph.D. student and researcher with the Institute
of Systems and Robotics and the Department of Electrical and Computer
Engineering, Faculty of Science and Technology, University of Coimbra.
His current research interests include the interplay of differential
geometry with computer vision and pattern recognition.
\end{IEEEbiographynophoto}

\begin{IEEEbiography}[{\includegraphics[clip,width=1in,height=1.25in,keepaspectratio]{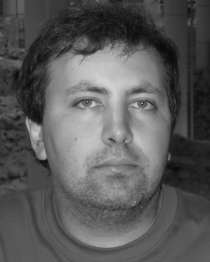}}]{Pedro Martins}
 received both his M.Sc. and Ph.D. degrees in Electrical Engineering
from the University of Coimbra, Portugal in 2008 and 2012, respectively.
Currently, he is a Postdoctoral researcher at the Institute of Systems
and Robotics (ISR) in the University of Coimbra, Portugal. His main
research interests include non-rigid image alignment, face tracking
and facial expression recognition.
\end{IEEEbiography}

\begin{IEEEbiography}[{\includegraphics[clip,width=1in,height=1.25in]{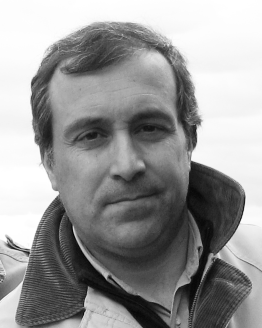}}]{Prof. Jorge Batista}
 received the M.Sc. and Ph.D. degree in Electrical Engineering from
the University of Coimbra in 1992 and 1999, respectively. He joins
the Department of Electrical Engineering and Computers, University
of Coimbra, Coimbra, Portugal, in 1987 as a research assistant where
he is currently an Associate Professor. Prof. Jorge Batista was the
Head of the Department of Electrical Engineering and Computer from
the Faculty of Science and Technology of University of Coimbra from
November 2011 until November 2013 and is a founding member of the
Institute of Systems and Robotics (ISR) in Coimbra, where he is a
Senior Researcher. His research interest focus on a wide range of
computer vision and pattern analysis related issues, including real-time
vision, video surveillance, video analysis, non-rigid modeling and
facial analysis. More recently his research activity also focus on
the interplay of Differential Geometry in computer vision and pattern
recognition problems. He has been involved in several national and
European research projects, several of them as PI at UC.
\end{IEEEbiography}

\vfill{}

\end{document}